\documentclass[letterpaper,11pt,leqno]{article}
\usepackage{paper}
\pdfoutput=1

\newcommand{\ie}{\emph{i.e.}}
\newcommand{\eg}{\emph{e.g.}}

\usepackage{wrapfig}
\usepackage[most]{tcolorbox}

\usepackage{setspace}
\usepackage{microtype}
\usepackage{hyperref}
\hypersetup{
    colorlinks=true, 
    linkcolor=darkblue,
    citecolor=darkblue,
    urlcolor=darkblue
}

\usepackage{tabularx} 
\usepackage{geometry} 
\usepackage{booktabs}

\usepackage{xcolor}
\definecolor{darkblue}{RGB}{0, 0, 100}
\newcommand{\myexample}[2]{
    \begin{tcolorbox}[colback=black!5!white,colframe=black,title={#1}]
        #2
    \end{tcolorbox}
}
\usepackage{graphicx} 

\usepackage[utf8]{inputenc}
\usepackage{xspace}

\usepackage{bbm}
\usepackage{cases}

\usepackage{fullpage}

\hypersetup{pdftitle={AI and the Problem of Knowledge Collapse}}

\begin{document}

\title{AI and the Problem of Knowledge Collapse}

\author{Andrew J. Peterson
\thanks{University of Poitiers. Under review. Replication code available on  \href{https://github.com/aristotle-tek/knowledge-collapse}{Github here}.}}

\date{April 2024}   

\begin{titlepage}
\maketitle

While artificial intelligence has the potential to process vast amounts of data, generate new insights, and unlock greater productivity, its widespread adoption may entail unforeseen consequences. We identify conditions under which AI, by reducing the cost of access to certain modes of knowledge,  can paradoxically harm public understanding. While large language models are trained on vast amounts of diverse data, they naturally generate output towards the `center' of the distribution.  This is generally useful, but widespread reliance on recursive AI systems could lead to a process we define as ``knowledge collapse'', and argue this could harm innovation and the richness of human understanding and culture. However, unlike AI models that cannot choose what data they are trained on, humans may strategically seek out diverse forms of knowledge if they perceive them to be worthwhile. To investigate this, we provide a simple model in which a community of learners or innovators choose to use traditional methods or to rely on a discounted AI-assisted process and identify conditions under which knowledge collapse occurs. In our default model, a 20\% discount on AI-generated content generates public beliefs 2.3 times further from the truth than when there is no discount. An empirical approach to measuring the distribution of LLM outputs is provided in theoretical terms and illustrated through a specific example comparing the diversity of outputs across different models and prompting styles. Finally, we consider further research directions to counteract harmful outcomes.

\end{titlepage}

\section{Introduction}
\label{Introduction}

Before the advent of generative AI, all text and artwork was produced by humans, in some cases aided by tools or computer systems.  The capability of large language models (LLMs) to generate text with near-zero human effort, 
however, along with models to generate images, audio, and video, suggest that the data to which humans are exposed may come to be dominated by AI-generated or AI-aided processes. 

Researchers have noted that the recursive training of AI models on synthetic text may lead to degeneration, known as ``model collapse'' \citep{shumailov2023curse}. Our interest is in the inverse of this concern, focusing instead on the equilibrium effects on the distribution of knowledge within human society.  We ask under what conditions the rise of AI-generated content and AI-mediated access to information might harm the future of human thought, information-seeking, and knowledge.

The initial effect of AI-generated information is presumably limited, and existing work on the harms of AI rightly focuses on the immediate effects of false information spread by ``deepfakes'' \citep{heidari2023deepfake}, bias in AI algorithms \citep{nazer_bias_2023}, and political misinformation \citep{chen_combating_2023}. Our focus has a somewhat longer time horizon, and probes the impact of widespread, rather than marginal adoption.

Researchers and engineers are currently building a variety of systems whereby AI would mediate our experience with other humans and with information sources. These range from learning from LLMs \citep{chen_artificial_2020}, ranking or summarizing search results with LLMs \citep{sharma_generative_2024}, suggesting search terms or words to write as with traditional autocomplete \citep{graham_ethical_2023,chonka_algorithmic_2023}, designing systems to pair collaborators \citep{ball_mass_2018}, LLM-based completion of knowledge bases sourced from Wikipedia \citep{chen_knowledge_2023}, interpreting government data \citep{fisher_uk_2024} and aiding journalists \citep{opdahl2023trustworthy}, to cite only a few from an ever-growing list.

Over time, dependence on these systems, and the existence of multifaceted interactions among them, may create a ``curse of recursion'' \citep{shumailov2023curse}, in which our access to the original diversity of human knowledge is increasingly mediated by a partial and increasingly narrow subset of views. With increasing integration of LLM-based systems, certain popular sources or beliefs which were common in the training data may come to be reinforced in the public mindset (and within the training data), while other ``long-tail'' ideas are neglected and eventually forgotten.

Such a process might be reinforced by an `echo chamber' or information cascade effect, in which repeated exposure to this restricted set of information leads individuals to believe that the neglected, unobserved tails of knowledge are of little value. To the extent AI can radically discount the cost of access to certain kinds of information, it may further generate harm through the ``streetlight effect'', in which a disproportionate amount of search is done under the lighted area not because it is more likely to contain one's keys but because it's easier to look there. We argue that the resulting curtailment of the tails of human knowledge would have significant effects on a range of concerns,  including fairness, inclusion of diversity, lost-gains in innovation, and the preservation of the heritage of human culture.

In our simulation model, however, we also consider the possibility that humans are strategic in actively curating their information sources. If, as we argue, there is significant value in the tai' areas of knowledge that come to be neglected by AI-generated content, some individuals may put in additional effort to realize the gains, \emph{assuming they are sufficiently informed} about the potential value.

\subsection{Summary of Main Contributions}

We identify a dynamic whereby AI, despite only reducing the cost of access to certain kinds of information, may lead to ``knowledge collapse,'' neglecting the long-tails of knowledge and creating an degenerately narrow perspective over generations. We provide a positive knowledge spillovers model with in which individuals decide whether to rely on cheaper AI technology or invest in samples from the full distribution of true knowledge. We examine through simulations the conditions under which individuals are sufficiently informed to prevent knowledge collapse within society. To evaluate this empirically, we outline an approach to defining and measuring output diversity, and provide an illustrative example. Finally, we conclude with an overview of possible solutions to prevent knowledge collapse in the AI-era.

\section{Previous Work}
\label{previous-work}

Technology has long affected how we access knowledge, raising concerns about its impact on the transmission and creation of knowledge. Yeh Meng-te, for example, argued in the twelfth century that the rise of books led to a decline in the practice of memorizing and collating texts that contributed to a decline of scholarship and the repetition of errors \citep{cherniack_book_1994}. 
Even earlier, a discussion in Plato's \emph{Phaedrus} considers whether the transition from oral tradition to reading texts was harmful to memory, reflection and wisdom \citep{hackforth1972plato}.
 
 We focus on recent work on the role of digital platforms and social interactions, and mention only in passing the literature on historical innovations and knowledge \citep{ong_orality_2013,mokyr_gifts_2011,havelock_literate_2019}, and the vast literature on the printing press  \citep{dittmar_information_2011,eisenstein_printing_1980}. Like other media transitions before it \citep{wu2011master}, the rise of internet search algorithms and of social media raised concerns  about the nature and distribution of information people are exposed to, and the downstream effects on attitudes and political polarization \citep{cinelli_echo_2021,barbera_social_2020}.

The following section considers research on the impact of recommendation algorithms and self-selection on social media, and how this might generate distorted and polarizing opinions, as an analogy for understanding the transformation brought about by reliance on AI. We consider game theoretic models of information cascades as an alternative model for failure in social learning, in which the public to fails to update rationally on individuals' private signals. Next, we review the main findings of network analysis on the flow of information in social media, which also identify mechanisms which distort knowledge formation. We then examine the specific nature of generative AI algorithms, focusing on the problem of model collapse and known biases in AI outputs.

\subsection{The media, filter bubbles and echo chambers} 

A common critique of social media is that they allow users to select in to ``echo chambers'' (specific communities or communication practices) in which they are exposed to only a narrow range of topics or perspectives.  For example, instead of consulting the ``mainstream'' news where a centrist and relatively balanced perspective is provided, users are exposed to selective content that echoes pre-existing beliefs.  In the ideological version of the echo-chamber hypothesis, individuals within a latent ideological space (for example a one-dimensional left-right spectrum), are exposed to peers and content with ideologically-similar views. If so, their beliefs are reinforced socially and by a generalization from their bounded observations, leading to political polarization \citep{cinus_effect_2022,jamieson_echo_2008,pariser2011filter}.
 
A simple model for this assumes homophily within in a network growth model, in which similar individuals chose to interact. Implicitly the approach presumes that this is common on social media but not common within traditional media, which for technological reasons were constrained to provide the same content across a broad population with possibly heterogeneous preferences.\footnote{The reality is as usual more complex. For example, in the post-war era, the concern was almost the inverse- the fear that the few channels that were possible with television led to `homogenization.'  There are also other dynamics at play than technological constraints. For example, in contrast to TV, the 1950s and 1960s saw a proliferation of more diverse and local radio stations, some catering to ethnic minorities and musical tastes outside the mainstream. The `payola' scandals, however, led to regulations that shifted content decisions from diverse DJs to centralized music directors \citep{douglas2002mass}.} 
This general dynamic may hold even if traditional media and newspapers were themselves dynamic systems interacting with their consumers, markets and advertisers, and themselves adapting their message to specific communities and preferences \citep{angeluccimedia,cage_media_2020,boone_resource_2002} .

The second main line of analysis focuses on ``filter bubbles,'' whereby the content to which users are exposed is selected based on a recommendation system. Jiang, et al., model this as a dynamic process between a user's evolving interests and behavior (such as clicking a link, video, or text) and a recommender system which aims to maximize expected utility for the user \citep{jiang_degenerate_2019}. In their reinforcement learning-inspired framework, the aim is for the user to explore the space of items or topics without the algorithm assigning degenerate (extremely high or zero) probabilities to these items. As above, a key concern is the political or ideological content of recommendations their relation to polarization \citep{keijzer_complex_2022}. In a more recent twist, \citep{sharma_generative_2024} find that LLM-powered search may generate more selective exposure bias and polarization by reinforcing pre-existing opinions based on finer-grained clues in the user's queries. 

Particularly relevant for our context is the issue of ``popularity bias'' in recommender systems, in which a small subset of content receives wide exposure while users (distributed based on some long-tailed distribution, like the topics) from smaller groups or with rare preferences are marginalized.  On the one hand, users may desire to be exposed to popular content, for example to understand trending ideas or fashions. But overly favoring popular items can lead to user disengagement because it neglects their unique interests, lacks variety, etc. \citep{klug2021trick} .  Recommendation systems are often biased in the sense that even when a subset of users \emph{wants} to get access to non-popular items, they receive few or no such recommendations. A number of approaches have been suggested to counteract this tendency \citep{lin_quantifying_2022,gao_cirs_2023}.

The problem of popularity bias is ironic given that one of the unique contributions of the internet was its ability to provide access to long-tailed products and services that were previously ignored or inaccessible \citep{brynjolfsson2006niches,brynjolfsson_consumer_2003}. By extension, we would expect social media and the internet to make possible a more diverse and rich informational environment. The role of self-selection into communities and recommendation algorithms provides a explanation for why this might not be the case.  In the next section we consider a more general set of models that examine information flow within networks and the idea of information cascades.

\subsection{Network effects and Information Cascades}

Information cascade models provide one approach to explaining a kind of herd behavior (where diverse and free individuals nonetheless make similar decisions). They explore the conditions under which private information is not efficiently aggregated by the public. This can occur where individuals sequentially make decisions from a discrete set after observing the behaviors but not the private signals of others. This can generate a ``herd externality'' \citep{banerjee_simple_1992} in which an individual ignores her private signal in deciding, and as a result the public is in turn unable to update on her private information. In the extreme, this can mean that \emph{all} private information, aside from that of the first few individuals, is completely ignored \citep{bikhchandani_learning_1998,smith_pathological_2000}.  In some variants of the model, individuals must pay to receive a signal, which encourages the tendency to want to free-ride on the information received by others, and thus the greater the cost, the more likely it is that a cascade develops.

A related literature on the spread of information on social networks analyzes information cascades in terms of network structure, as a kind of contagion. Here, the focus is not on private information but how information flows within the network. For example, independent cascade models consider how an individual may change their beliefs based on some diffusion probability as a result of contact with a neighbor with that belief \citep{goldenberg_talk_2001,gruhl_information_2004}.

More generally, such models determine the probability of diffusion within a network as some function of the connected nodes, and may also incorporate additional characteristics such as each nodes' social influence, ideological or other preferences, or topics \citep{barbieri_topic-aware_2013}. Alternatively, epidemic models allow that individuals may be in one of three states - susceptible, infected (capable of transmitting the information), and recovered (in which case they have the information but do not consider it worth sharing with others) \citep{kermack1927contribution,barrat_dynamical_2008} . 

Social (and even physical) proximity can lead individuals to share similar attitudes, such as when individuals randomly assigned housing together come to have attitudes similar to their apartment block and differing from nearby blocks  \citep{festinger_social_1950,nowak_private_1990}. Empirically, some claim that weak-ties may be more important for information diffusion that strong-ties \citep{bakshy_role_2012}, while another approach focuses on clustering within the network as a means for spreading information \citep{centola_spread_2010}. More sophisticated models allow for the evolution not only of opinion process but the edges between nodes of the network \citep{castellano2009statistical}. 

These models suggest specific opinion-formation dynamics based on what other humans, texts, images, etc. an individual interacts with. By extension, we could consider the generalization of these networks to the case where LLMs play a key role as (possibly influential) nodes, or as determining how an individual navigates a knowledge graph. One of the key ideas of Web 2.0 was that users, not just authors or programmers, structure the knowledge \citep{oreilly_web20_2005}. By extension, in the AI era, LLMs interact with users, authors, programmers and technology to structure that knowledge, and understanding the flow of information requires understanding the emergent behavior of these elements.

 \subsection{Model collapse}

The idea of model collapse is rooted in the earlier phenomenon of ``mode collapse'' in generative adversarial networks (GANs). GANs are based on a generator neural network that proposes, e.g. an image, and a discriminator attempts to predict whether a given image is created by the generator or is a real image from the dataset. While ideally the generator attempts to produce images across the full range of input data, in practice they may settle into producing a narrow range of images for which it is good at fooling the discriminator, known as mode collapse  \citep{goodfellow2016nips,arora_generalization_2017}. The case of ``posterior collapse'' was also identified in modeling language data with variational autoencoders \citep{melis_mutual_2022}.

Shumailov, et al. (2023) introduced the term ``model collapse'' to describe a related process when models such as variational autoencoders, Gaussian mixture models, and LLMs are trained on data produced by an earlier version of the model. Incorporating AI-generated content in the training data causes loss of information which they categorize into two types. First, in ``early model collapse,'' the tails of the distribution are lost due to statistical error (finite sampling bias) or functional approximation error, which leads to reversion to the mean.  Second, ``late model collapse'' may occur when a model converges with narrow variance on a distribution unlike the original data. They provide evidence of such model collapse in LLMs and other models, see for example Figure~\ref{fig:shumailov}.

\begin{figure}[ht]
\centering 
\myexample{ \footnotesize (Reproduced from Shumailov et al. \citeyear{shumailov2023curse})\\
Example of text outputs of an OPT-125m model affected by~\emph{Model Collapse} -- models degrade over generations, where each new generation is trained on data produced by the previous generation.}{
\begin{description} \footnotesize
    \item \textbf{Input:} some started before 1360 — was typically accomplished by a master mason and a small team of itinerant masons, supplemented by local parish labourers, according to Poyntz Wright. But other authors reject this model, suggesting instead that leading architects designed the parish church towers based on early examples of Perpendicular
    \item \textbf{Outputs:}
    
    \item \textbf{~Gen 0:} Revival architecture such as St. John's Cathedral in London. The earliest surviving example of Perpendicular Revival architecture is found in the 18th @-@ century Church of Our Lady of Guernsey, which dates from the late 19th century. There are two types of perpendicular churches : those

    \item \textbf{~Gen 1:}  architecture such as St. Peter's Basilica in Rome or St. Peter's Basilica in Buenos Aires. There is no evidence that any of these buildings were built during the reign of Pope Innocent III, but it is possible that they may have been built during the reign of his successor, Pope Innocent

    \item \textbf{~Gen 7:}  architecture in England. In an interview with The New York Times, Wright said : " I don 't think there is anything wrong with me being able to do what I want to do. It just doesn 't work for me. " He added : " I don 't know if you can call it funny,

    \item \textbf{~Gen 9:} architecture. In addition to being home to some of the world's largest populations of black @-@ tailed jackrabbits, white @-@ tailed jackrabbits, blue @-@ tailed jackrabbits, red @-@ tailed jackrabbits, yellow @-
\end{description} \normalsize
}
\caption{Model collapse example from Shumailov 2023.}
\label{fig:shumailov}
\end{figure}

On the one hand, under some conditions the injection of true (non AI-generated) data can preserve representation of the true distribution \citep{dohmatob_tale_2024}, though in other models, small amounts of synthetic data can poison an image model, and once distorted, it is difficult for such models to recover even after being trained on true data \citep{bohacek_nepotistically_2023}. As one particular example, training LLMs on synthetic data can lead to diminishing lexical, semantic and syntactic diversity \citep{guo_curious_2023}.

\subsection{Known biases in LLMs}

Newer AI models such as LLMs are not immune to the problems of bias identified and measured in machine learning algorithms \citep{nazer_bias_2023} and which have plagued predictive algorithms in real-world uses cases going back to at least the 1930s \citep{christian2021alignment}. Unsurprisingly, LLMs are better at recalling facts that occur frequently within the training data and struggle with long-tail knowledge \citep{kandpal_large_2023}.  \citep{das_under_2024} identify a range of shortcomings of LLMs in attempting to generate human-like texts, such as under-representing minority viewpoints and reducing the broad concept of ``positive'' text to that simply of expressing ``joy''.

Recent work attempts to address these issues through a variety of methods, for example by upsampling under-represented features on which prediction is otherwise sub-optimal \citep{gesi_leveraging_2023}, or evaluating the importance of input data using shapely values \citep{karlas_data_2022}. However, the mechanistic interpretability work on LLMs to date suggest that our understanding, while improving, is still very limited \citep{kramar_atp_2024,wu_interpretability_2023}. As such, direct methods for overcoming such biases are, at a minimum, not close at hand. Finally, while much of the focus is naturally on overt racial and gender biases, there may also be pervasive but less observable biases in the content and form of the output. For example, current LLMs trained on large amounts of English text may `rely on' English in their latent representations, as if a kind of reference language \citep{wendler_llamas_2024}.

One particular area in which the diversity of LLM outputs has been analyzed is on a token-by-token level in the context of decoding strategies. In some situations, using beam search to choose the most likely next token can create degenerate repetitive phrases \citep{su_contrastive_2022}.  Furthermore, a bit like Thelonious Monk's melodic lines, humans do not string together sequences of the most likely words but occasionally try to surprise the listener by sampling from low-probability words, defying conventions, etc. \citep{holtzman_curious_2020}.

\section{A Model of Knowledge Collapse}
\label{sec:model}

\subsection{Defining Knowledge Collapse}

A commonly held, optimistic view is that knowledge has improved monotonically over time, and will continue to do so. This indeed appears to be the case for certain scientific fields like physics, chemistry, or molecular biology, where we can measure the quality of predictions made over time. For example, accuracy in the computation of digits of $\pi$ has increased from 1 digit in 200 BCE to 16 in 1424 (Jamashid al-Kashi) to $10^{14}$ digits recently.

In other domains, however, it is less clear, especially within regions. Historically, knowledge has not progressed monotonically, as evidenced by the fall of the Western Roman empire, the destruction of the House of Wisdom in Baghdad and subsequent decline of the Abbasid Empire after 1258, or the collapse of the Mayan civilization in the 8th or 9th century.  Or, to cite specific examples, the ancient Romans had a recipe for concrete that was subsequently lost, and despite progress we have not yet re-discovered the secrets of its durability \citep{seymour_hot_2023}, and similarly for Damascus steel \citep{kurnsteiner_high-strength_2020}. Culturally, there are many languages, cultural and artistic practices, and religious beliefs that were once held by communities of humans which are now lost in that they do not exist among any known sources \citep{nettle_vanishing_2000}.

The distribution of knowledge across individuals also varies over time. For example, traditional hunter-gatherers could identify thousands of different plants and knew their medicinal usages, whereas most humans today only know a few dozen plants and whether they can be purchased in a grocery store. This could be seen as a more efficient form of specialization of information across individuals, but it might also impact our beliefs about the value of those species or of a walk through a forest, or influence scientific or policy-relevant judgements.

Informally,\footnote{For further discussion and a more precise definition using the notation from the model, see the Appendix.} we define knowledge collapse as the progressive narrowing over time (or over technological representations) of the set of information available to humans, along with a concomitant narrowing in the perceived availability and utility of different sets of information. The latter is important because for many purposes it is not sufficient for their to exist a \emph{capability} to, for example, go to an archive to look up some information. If all members deem it too costly or not worthwhile to seek out some information, that theoretically available information is neglected and useless.

\subsection{Model Overview}

The main focus of the model is whether individuals decide to invest in innovation or learning (we treat these as interchangeable) in the `traditional' way, through a possibly cheaper AI-enabled process, or not at all.  The idea is to capture, for example, the difference between someone who does extensive research in an archive rather than just relying on readily-available materials, or someone who takes the time to read a full book rather than reading a two-paragraph LLM-generated summary.

Humans, unlike LLMs trained by researchers, have agency in deciding among possible inputs. Thus, a key dynamic of the model is to allow for the possibility that rational agents may be able to prevent or to correct for distortion from over-dependence on `centrist' information. If past samples neglect the `tail' regions, the returns from such knowledge should be relatively higher. To the extent that they observe this, individuals would be willing to pay more (put in more labor) to profit from these additional gains. We thus investigate under what conditions such updating among individuals is sufficient to preserve an accurate vision of the truth for the community as a whole.

The cost-benefit decision to invest in new information depends on the expected value of that information. Anyone who experiments with AI for, e.g. text summarization, develops an intuitive sense of when the AI provides the main idea sufficiently well for a given purpose and when it is worth going straight to the source. We assume that individuals cannot foresee the future, but they do observe in common the realized rewards from previous rounds. The decision also depends on each individual's type. Specifically, $n$ individuals have types $\Theta_n$ drawn from a lognormal distribution with $\mu=1$, $\sigma=0.5$. Depending on how their utility is calculated (not a substantive focus here), these could be interpreted as different expected returns from innovation (\eg techno-optimists versus pessimists), or their relative ability or desire to engage in innovation.

\begin{table}[ht]
\centering \footnotesize
\caption{Summary of notation}
\begin{tabular}{ll}
\hline
\textbf{Notation} & \textbf{Description} \\ \hline
$n$           & number of individuals ($=25$)     \\
$\Theta_n$ & the type of individual $n$, multiplying their expected return from innovation \\
$d.f.$           & degrees of freedom for t-distribution, determines width of the tails   \\
$p_{\text{true}}(x)$       & the `true' probability distribution function, t-distribution with e.g. 10 d.f. \\
$p_\text{public} (x)$     & the public approximation to the true pdf based on the 100 most \\
                                     &  recent samples using kernel density estimation   \\
$\delta$ & AI-discount rate, where the cost of an AI-sample is $\delta$ times \\ 
                                     &  the cost of a sample from the full distribution \\
$\sigma_{tr}$          & truncation limits for the AI-generated samples, in standard deviations   \\
$\hat{v}_{t}$         & The estimated value of a sample at time $t$ \\
$\eta$           & learning rate, i.e. how quickly individuals update \\
                 & their beliefs on the value of full and truncated \\
                 & samples based on samples observed in the last round   \\
$r$           & How many rounds between generations   \\
                                     & (if greater than 100, no generational effects) \\
$\mathbb{I}$           & Innovation from an individual's sample \\
     & (i.e. how far they move the public pdf towards the true pdf) \\
     & determines the $n$'s payout when multiplied by $\Theta_n$  \\
\end{tabular}
\end{table}
 
We model knowledge as a process of approximating a (Students t) probability distribution.\footnote{Full replication code available at:\\ https://github.com/aristotle-tek/knowledge-collapse } This is simply a metaphor, although it has parallels for example in the analysis of model collapse \citep{shumailov2023curse}, but we make no claim that ``truth'' is in some deep way distributed 1-D Gaussian. This is a modeling assumption in order to work with a process with well-known properties, where there is both a large central mass and long-tails, which we take to be in some general way reflective of the nature of knowledge (and of the distribution of training data for LLMs.)

The set of individuals who decide to invest in information receive a sample from the true distribution,  while those that invest in the AI-generated sample receive a sample from a version of the true distribution which is truncated at $\sigma_{tr}$ standard deviations above and below the mean. To vary the extent of mass in the tails, we model the true distribution as a Student's t-distribution with \eg~$10$ degrees of freedom. The results are similar for a standard normal distribution, and as expected the problem of knowledge collapse is more pronounced for wider tails (c.f. Appendix Figure~\ref{fig:tail_thickness}).

While individuals choose whether or not to invest in innovation according to their personal payoff, when they do so invest they also contribute their knowledge to the public. That is, a \emph{public} knowledge probability distribution function (`public pdf') is generated by gathering the $nsamp=100$ most recent samples\footnote{Varying this has trivial effect on the model, though higher values can distort public knowledge.}  and generating an estimate of the truth using kernel density estimation. The distance between the public pdf and the truth provides a shorthand for the general welfare of a society. We define knowledge collapse as occurring where there is a large and increasing distance between the public and true pdfs as a result of the collapse of tail regions and increasing mass near the mean.

The individual's payoff is calculated according to the distance they move the public pdf towards the true pdf. That is, the innovation (individual payoff) $\mathbb{I}$ generated by an individuals additional ($n+1$)th sample is calculated with respect to the true pdf $p_\text{true}(x)$ and the current public pdf $p_\text{public} (x)$, based on the Hellinger distance $H(p(x), q(x))$\footnote{We use the Hellinger distance because it is a true distance metric that is symmetric and satisfies the triangle inequality, which is important for the innovation calculation. The Hellinger distance is bounded by 0  and 1 (if the two pdfs have no common support) and given by: \\
$ H(p, q) = \frac{1}{\sqrt{2}} \sqrt{\int \big(\sqrt{p(x)} - \sqrt{q(x)} \big) ^2 dx} $
}, as follows:
\[
\text{innovation} =   \text{previous distance} - \text{new distance}
\] 
\[
\mathbb{I} =   H \Big(p^{n}_{\text{public}} (x) - p_{\text{true}}(x)  \Big) - H \Big( p^{n+1}_{\text{public}} (x),  p_{\text{true}}(x) \Big) 
\]

\begin{figure}[!ht] 
  \begin{center}
\begin{tikzpicture}[scale=1]
  \node (pn1) at (0,0) [label=below:$p^{n+1}_{\text{public}}$] {};
  \node (ptrue) at (4,0) [label=below:$p_{\text{true}}$] {};
  \node (pn) at (0,3) [label=above:$p^{n}_{\text{public}}$] {};

  \draw (pn1) -- node[below] {0.4} (ptrue);
  \draw (ptrue) -- node[right] {0.5} (pn);
  
  \draw[->, dashed] (pn) -- node[left, rotate=90, yshift=6mm, xshift=12mm] {$\Delta$ new sample} (pn1);

  \fill (pn) circle (2pt);
  \fill (pn1) circle (2pt);
  \fill (ptrue) circle (2pt);
  
  \node at (2, -1) {$\mathbb{I} = 0.5 - 0.4 = 0.1$};
\end{tikzpicture}

  \end{center}
  \caption{A hypothetical innovation calculation where the new (n+1)th sample moves the public pdf $0.1$ towards the true distribution.}
   \vspace{10pt}
  \label{fig:dist_ex}
\end{figure}
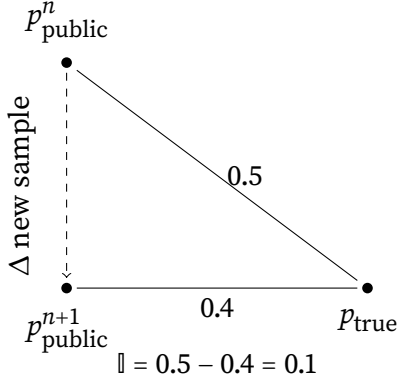

In Figure~\ref{fig:dist_ex}, we illustrate the innovation calculation for a hypothetical example where the distance between the existing public pdf and the true pdf is $0.5$, while the $n+1$th sample reduces the distance to $0.4$, thereby generating an innovation of $0.1$.

\noindent This can be thought of as akin to a patent process, in which an individual receives rents for her patent (to the extent that it is truly innovative) in exchange for contributing to public knowledge that benefits others. 

As noted above, individuals cannot foresee the true future value of their innovation options (they do not know what sample they will receive or how much value it will add. Instead, they can only estimate the relative values of innovation based on the previous rounds. Specifically, they update their belief about the options based on the previous full and truncated (AI) samples from the previous round (and a minimum of three), according to a learning rate ($\eta$) as follows. For the previous estimate $\hat{v}_{t-1}$, the new estimate $\hat{v}_{t}$ for each of the full- and truncated-samples is calculated from the observed value in the previous round ($\mathbb{I}_{t-1}$) as:
\[
\hat{v}_{t} = \hat{v}_{t-1} + \eta \cdot \Big( \hat{v}_{t-1} - \mathbb{I}_{t-1} \Big)
\] 

\noindent By varying the learning rate, we can evaluate the impact of having more or less up-to-date information on the value of different information sources, where we expect that if individuals are sufficiently informed, they will avoid knowledge collapse by seeing and acting on the potential to exploit knowledge from the tail regions, even if relatively more expensive.

While the individual payoff is based on the true movement of the public pdf towards the true pdf, the public pdf is updated based on \emph{all} samples. This reflects that public consciousness is overwhelmed with knowledge claims and cannot evaluate each, so that a consensus is formed around the sum of all voices. Unlike the individual innovator who has a narrow focus and observes whether her patent ultimately generates value, the public sphere has limited attention and is forced to accept the aggregate contributions of the marketplace of ideas.

As a result, individuals' investments in innovation have positive spillovers to the extent they can move public knowledge towards the truth. However, if too many people invest in `popular' or `central' knowledge by sampling from the truncated distribution, this can have a negative externality, by distorting public knowledge towards the center and thinning the tails.\footnote{If individuals knew they were sampling from a truncated distribution, they could use the Expectation-Maximization algorithm to recover the full distribution, but again this process is meant to be metaphorical, and there is no known real-life method for recovering the source knowledge from AI-generated content. }

We also introduce the possibility of generational turnover in some models to explore the impact on knowledge collapse. This could either be taken to be literal generations of humans, as in economic `overlapping generation' models \citep{weil_overlapping_2008}, or alternatively as reflecting the recursive nature of reliance on interleaved AI-systems, which could generate the same result within a rapid timeframe. 

In the version of the model with generational change, the new generation takes the existing public pdf to be representative and thus begins sampling from a distribution with the same (possibly smaller) variance (and correspondingly the truncation limits are updated). Interpreted in terms of human generations, this could be understood as the new generation fixing its `epistemic horizon' based on the previous generation. That is, the new generation may underestimate the breadth of possible knowledge and then rely on these perceived limits to restrict their search.\footnote{Zamora (2010) suggests a scientific process of `verisimiltude', where we judge evidence not with reference to objective truth by by ``perceived closeness to what we empirically know about the truth, weighted by the perceived amount of information this empirical knowledge contains'' \citep{zamora-bonilla_what_2010} 
. Work on human cultural transmission attempts to explain, for example, how the Tasmanians lost a number of useful technologies over time. \citep{mesoudi_multiple_2008,henrich_2004}} An information cascade model could justify such a situation if individuals assume that previous actors would have invested in tail knowledge if was valuable, and thus take the absence of such information as implying that it must be of little value.\footnote{For example, Christian communities at times actively promoted and preserved `canonical' texts while neglecting or banning others, with the result that those excluded from reproduction by scribes were taken to have little value. Perhaps the heliocentric view espoused by Aristarchus of Samos in the 3rd century BCE would have been more readily (re)considered if his works had not been neglected \citep{russo2003forgotten}. A number of authors, such as Basilides, are known to us today only through texts denouncing (and sometimes misrepresenting) their views \citep{layton1989significance}. }

A second interpretation views these `generations' not in terms of human populations but as a result of recursive dynamics among AI systems, such as when a user reads an AI-generated summary of an AI-written research article which was itself constructed from Wikipedia articles edited with AI, etc., a fancy version of the telephone game.

\section{Results}

\label{results}

\begin{figure}[!ht] 
  \begin{center}
        \includegraphics[width=0.48\textwidth]{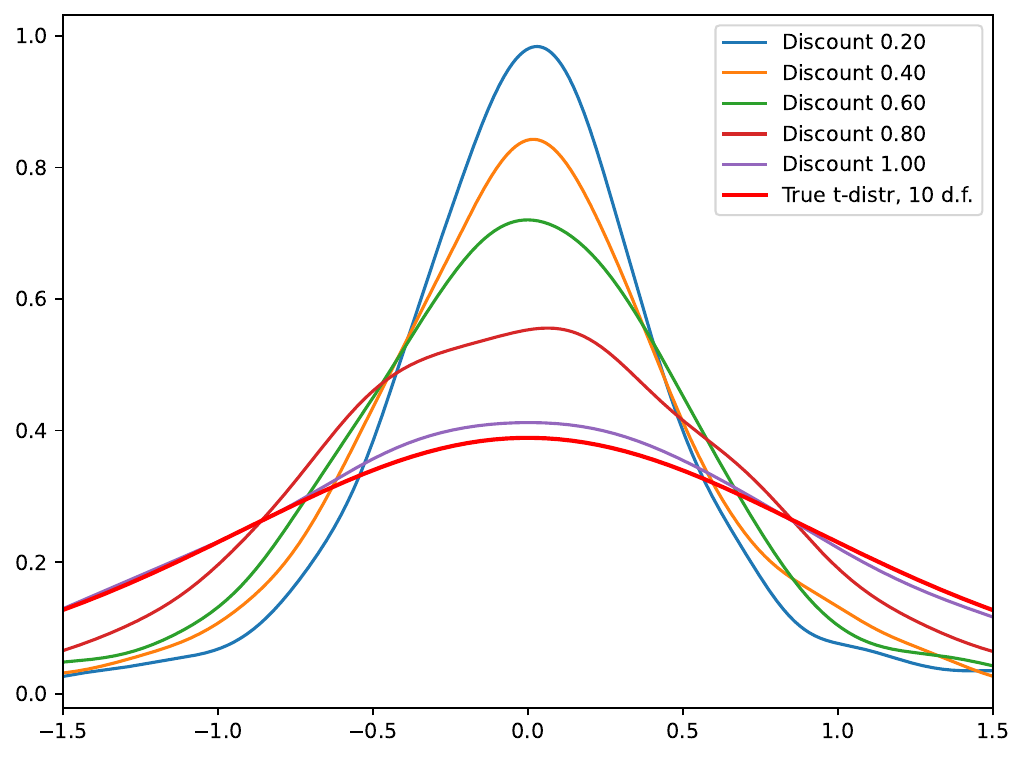}  
  \end{center}
  \caption{Knowledge collapse: The cheaper it is to rely on AI-generated content, the more extreme the degeneration of public knowledge towards the center.}
   \vspace{10pt}
  \label{fig:kde_discounts}
\end{figure}

Our main concern is with the view that AI, by reducing the costs of access to certain kinds of information, could only make us better off. In contrast to the literature on model collapse, we consider the conditions under which strategic humans may seek out the input data that will maintain the full distribution of knowledge. Thus, we begin with a consideration of different discount rates. First, we present the a kernel density estimate of public knowledge at the end of 100 rounds (Figure~\ref{fig:kde_discounts}). As a baseline, when there is no discount from using AI (discount rate is $1$), then as expected public knowledge converges to the true distribution,\footnote{Even with no discount, there are occasional samples from the truncated distribution, but only enough to realize that they are of relatively less worth than full-distribution samples} As AI reduces the cost of truncated knowledge, however, the distribution of public knowledge collapses towards the center, with tail knowledge being under-represented. Under these conditions, excessive reliance on AI-generated content over time leads to a curtailing of the eccentric and rare viewpoints that maintain a comprehensive vision of the world.

Fixing specific parameters, we can get a sense of the size of the the impact of relying on AI. For example, for our default model,\footnote{Truncation at $\sigma_tr=0.75$ standard deviations from the mean, generations every 10 rounds, learning rate of $0.05$.} after nine generations, when there is no AI discount the public distribution has a Hellinger distance of just $0.09$ from the true distribution\footnote{Even here there are occasional samples from the truncated distribution -- just enough to realize that they have less value than the full-distribution samples.}. When AI-generated content is 20\% cheaper (discount rate is $0.8$), the distance increases to $0.22$, while a 50\% discount increases the distance to $0.40$. Thus, while the availability of cheap AI-approximations might be thought to only increase public knowledge, under these conditions public knowledge is $2.3$ or $3.2$ times further away from the truth due to reliance on AI.

\begin{figure}
  \begin{center}
        \includegraphics[width=0.48\textwidth]{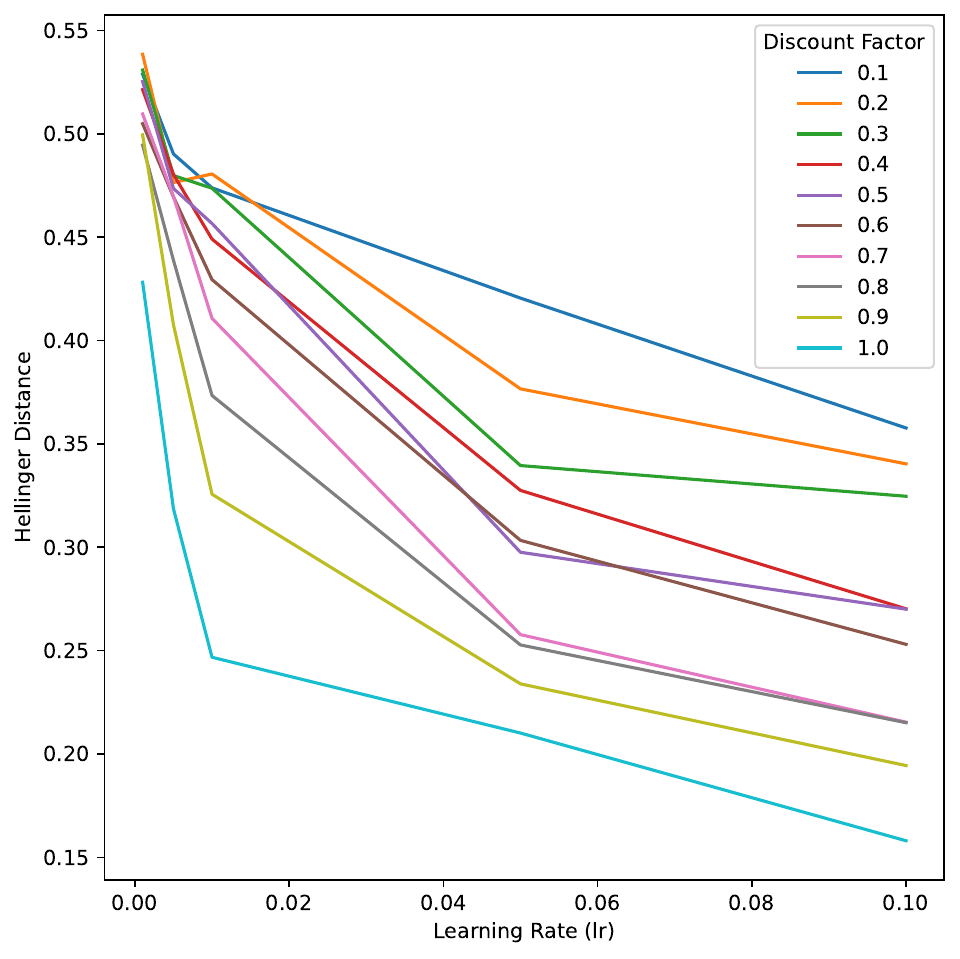}  
  \end{center}
  \caption{Discount rate and learning rate}
  \label{fig:dist_disc_lr}
     \vspace{10pt}
\end{figure}

For subsequent results illustrating the tradeoff of different parameters, we plot the Hellinger distance between public knowledge at the end of the 100 rounds and the true distribution. First, we examine the importance of updating on the value of relative samples and the relationship to the discount factor in Figure~\ref{fig:dist_disc_lr}. That is, we  compare the situation in which individuals do not update on the value of innovation in previous rounds (learning rate near zero, e.g. $\eta=0.001$) to the case where they update rapidly (here $\eta=0.1$). As above, the more AI-generated content is cheaper (discount rate indicated by colors), the more public knowledge collapses towards the center. At the same time, when individuals update more slowly on the relative value of learning from AI (the further to the left in the figure), the more public knowledge collapses. We also observe a tradeoff, that is, faster updating on the relative value of AI-generated content can compensate for more extreme price disparities. And conversely, if the discount rate is not too extreme, even slower updating on the relative values is not too harmful.

\begin{figure}
  \begin{center}
        \includegraphics[width=0.48\textwidth]{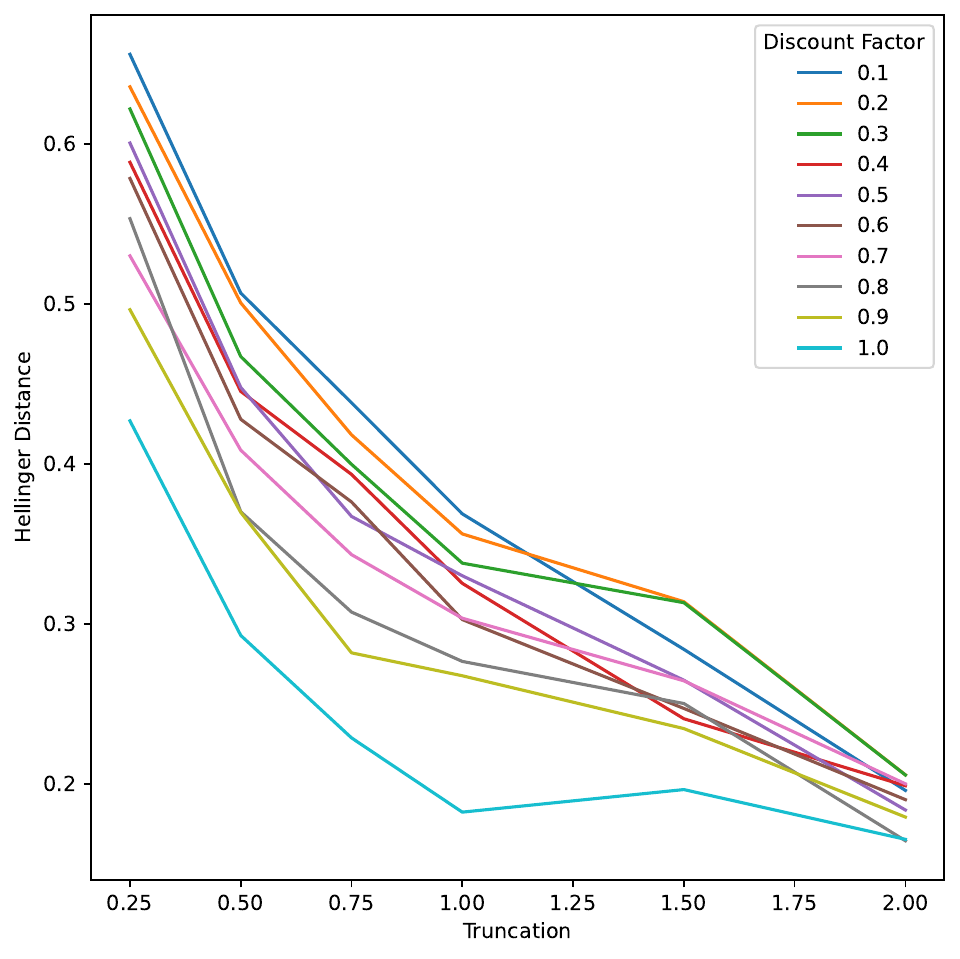}  
  \end{center}
  \caption{Discount rate and truncation limits} 
   \label{fig:dist_disc_sigmatr}
      \vspace{10pt}
\end{figure}

In Figure~\ref{fig:dist_disc_sigmatr}, we consider the impact of variations in how extreme the truncation of AI-generated content is on the collapse of knowledge. Intuitively, extreme truncation (small values of $\sigma_{tr}$) correspond to a situation in which AI, for example, summaries an idea with only the most obvious or common perspective. Less extreme truncation corresponds to the idea that AI manages to represent a variety of perspectives, and excludes only extremely rare or arcane perspectives. Naturally, in the latter case, (e.g. if AI truncates the distribution two standard deviations from the mean), the effect is minimal. If AI truncates knowledge outside of $0.25$ standard deviations from the mean, the impact is large, though once again this is at least someone moderated when the discount is smaller (especially if there is no generational effect).

\begin{figure}
  \begin{center}
        \includegraphics[width=0.48\textwidth]{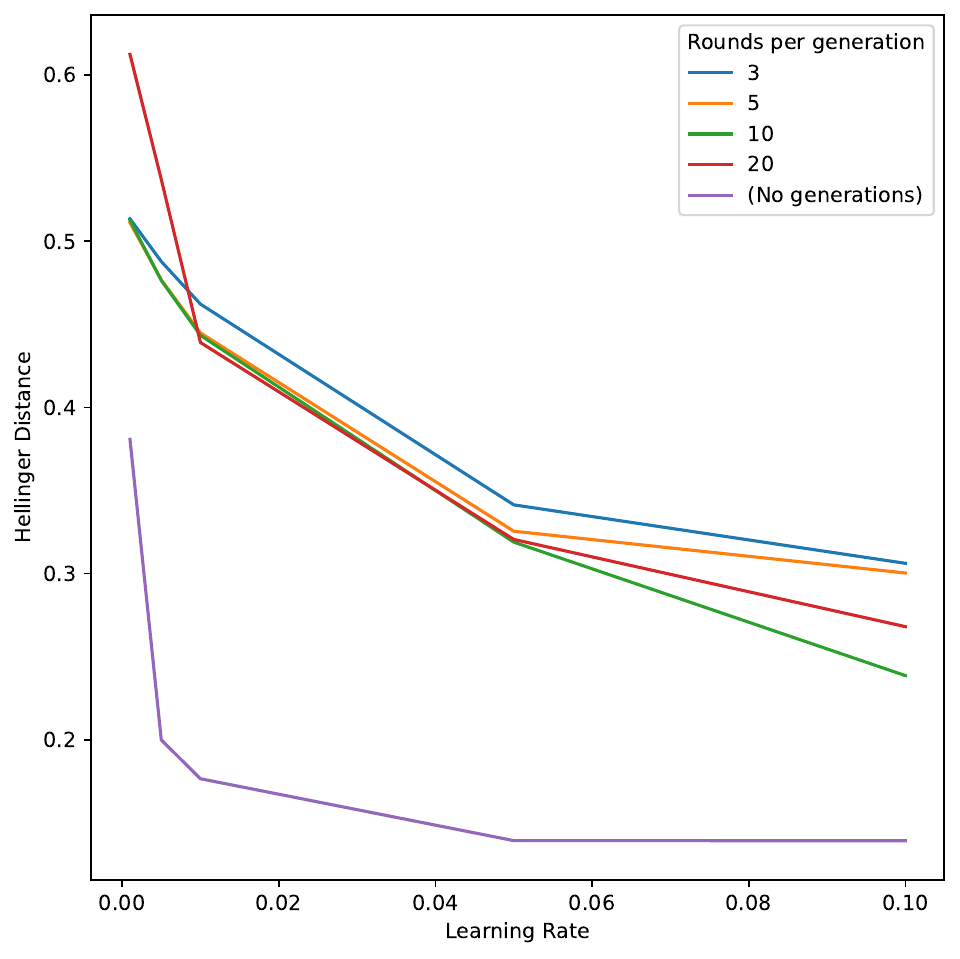}  
  \end{center}
  \caption{Learning rate and generational change}
  \label{fig:dist_lr_nrounds}
     \vspace{10pt}
\end{figure}

We compare the effect of the generational compounding of errors in Figure~\ref{fig:dist_lr_nrounds}. If there is no generational change, there is at worst only a reduction in the tails of public knowledge outside the truncation limits. In this case the distribution is stable and does not ``collapse'', that is, over time the problem is not progressively worse. We see a jump from this baseline to the case where there is generational change, though the effect of how often generational change occurs (every 3, 5, 10, or 20 rounds) does not have a significant impact. 

\section{Empirical Measurement of Diversity}
\label{empirical}

While the above discussion focused on the effects of medium- to long-term adoption, we provide a brief theoretical overview of approaches to measuring knowledge collapse. We also illustrate measuring the \emph{distribution} of LLM output in line with a specific empirical example, and provide a brief analysis of the extent to which direct prompting can help generate more diverse outputs. 
Unlike typical evaluations of machine learning algorithms, our focus is not on an evaluation metric relative to a set of gold-standard input- output pairs $x$ and $y$, but the diversity of outputs across a distribution of possible responses for a given question or prompt. For our illustration, we consider the question of ``What does the well-being of a human depend on?'', as a well-known question on which there no general consensus on a particular answer but instead every culture generally provides an answer in more or less explicit terms.

\subsection{Measuring Response Diversity}

We consider situations in which an LLM generates as outputs a set of items out of some possibly-larger set of possibilities. For example, we might ask the LLM to (repeatedly) produce a list of 100 human languages, poets, animal species, or chemical compounds. 

The LLM generates this set based on a transformer, state-space model, etc. $\mathcal{M}$, based on a set of parameters $\theta$ by generating, e.g. an auto-regressive set of tokens. A strong definition of representativeness 
would say that the probability of inclusion of any member $x_i$ in the output is equal to the probability of inclusion within a random sample from the full set $X$.

\vspace{1em}
\begin{definition}[\textsc{Strong Uniform Representativeness}]
Let $X = \{x_1, x_2, \ldots, x_n\}$ represent a finite set of possible items, and let $\mathcal{M}:\Theta \rightarrow \Delta(X)$ be a transformation governed by a set of parameters $\theta \in \Theta$, which maps to a distribution over the items in $X$. 

\emph{Strong Uniform Fairness} requires that the generation of items from the set $X$ by the model $\mathcal{M}$, for any given parameters $\theta$, should be uniformly distributed, such that
\begin{align}
    \forall x_i, x_j \in X. \quad P_{\mathcal{M}(\theta)}(x_i) = P_{\mathcal{M}(\theta)}(x_j) = \frac{1}{|X|}
\end{align}
where $P_{\mathcal{M}(\theta)}(x_i)$ denotes the probability of item $x_i$ being included in the output distribution generated by $\mathcal{M}$ with parameters $\theta$, and $|X|$ is the cardinality of set $X$.

\end{definition}
\vspace{1em}

\noindent While useful in some particular cases, or in the absence of alternatives, this is unreasonable for many real-world use cases. The other extreme is to introduce a simple non-degeneracy condition, saying that any item should have at least \emph{some} chance of appearing in the LLM output:

\vspace{1em}
\begin{definition}[\textsc{Minimal Representativeness}]

\emph{Minimal Representativeness} requires that for any item $x_i \in X$, the probability of $x_i$ being included in the output distribution generated by $\mathcal{M}$ with any given parameters $\theta$, is non-zero, such that
\begin{align}
    \forall x_i \in X. \quad P_{\mathcal{M}(\theta)}(x_i) > 0
\end{align}
where $P_{\mathcal{M}(\theta)}(x_i)$ denotes the probability of item $x_i$ being included in the output distribution generated by $\mathcal{M}$.
\end{definition}
\vspace{1em}

Intuitively, if asking an LLM for a diverse list of items, any possible item should have some non-zero probability of inclusion, analogous to the definition of non-degeneracy in a recommendation algorithm \citep{jiang_degenerate_2019}.

In some cases another possible practical approach is to define representativeness relative to a specific task. This implicitly defines a set of values across the different outcomes. For example, we might want the likelihood of a language of appearing in the output to be proportional to the number of currently-living speakers of that language. For some purposes, this would be the most helpful output, although in other cases, such as for a historian, it might be problematic because it fails to consider languages with no living speakers. Other users, for example, might want to consider the possibility of up-weighting at-risk languages in hopes of preserving them, etc.

\vspace{1em}

\begin{definition}[\textsc{ Task-Pragmatic Representativeness }]
Let $w:X \rightarrow \mathbb{R}_{\geq 0}$ be a weighting function that assigns a non-negative weight to each item $x_i \in X$, reflective of a specific covariate relevant to the task. 
\emph{Task-Pragmatic Representativeness} requires that the probability of any item $x_i$ being included in the output distribution generated by $\mathcal{M}$ with any given parameters $\theta$, is proportional to its weight, such that
\begin{align}
    P_{\mathcal{M}(\theta)}(x_i) = \frac{w(x_i)}{\sum_{x_j \in X} w(x_j)}
\end{align}
where $P_{\mathcal{M}(\theta)}(x_i)$ denotes the probability of item $x_i$ being included in the output distribution generated by $\mathcal{M}$.
\end{definition}

\noindent For example, a doctor asking for a list of possible infectious diseases to diagnose a patient might expect that it would begin with the most commonly occurring diseases first, before mentioning more obscure ones. Even this list depends on context, as there are certain infectious diseases are common worldwide (e.g. tuberculosis) but are rare in certain countries.

\vspace{1em}

Finally, a similar measure considers the extent to which different (e.g. protected) groups are represented in the output. As an example, in a small corpus created by asking Phi-2 2.7B and Llama2-13b to list twenty languages 100 times each, there is only one single mention of a non-verbal (Sign) language (``German Sign Language''), despite there being around 72 million people worldwide who speak sign languages. By contrast, Italian is spoken by around 68 million people, which is slightly less, but Italian was mentioned 508 times.  There are obvious reasons for this, in terms of the frequency of appearance of references to Italian in the text, but it is worth considering the possibility that this might reinforce biases and narrowness in thinking about languages. An objection might be that Italian is one language while the speakers of Sign languages use more than 300 different languages. Still, it's worth noting that Italian is 29th in the world in terms of number of speakers, so the list of twenty was not representative of the top twenty languages by speakers either, neglecting Tagalog 64 mentions; 83 million speakers, Tamil (58 mentions; 87 million speakers), etc. This suggests the value of a general measure of this:

\begin{definition}[\textsc{Group Proportional Representativeness}]
Let $P$ be a population partitioned into groups $\{G_1, G_2, \ldots, G_k\}$, where each group $G_i$ encompasses a set of individuals sharing specific characteristics (e.g., ethnicity, gender).
Let $w: \{G_1, G_2, \ldots, G_k\} \rightarrow \mathbb{R}_{\geq 0}$ be a weighting function that assigns a non-negative weight to each group $G_i$, based on a relevant metric (such as the proportion of the population).

\emph{Group Proportional Representativeness} requires that the probability of any individual from group $G_i$ being included in the output generated by $\mathcal{M}(\theta)$, is proportional to the weight of the group, such that
\begin{align}
    P_{\mathcal{M}(\theta)}(x \in G_i) = \frac{w(G_i)}{\sum_{j=1}^k w(G_j)}
\end{align}
for all $x \in G_i$ and for each group $G_i$, where $P_{\mathcal{M}(\theta)}(x \in G_i)$ denotes the probability of an individual from group $G_i$ being included in the output distribution generated by $\mathcal{M}$.

\end{definition}

\vspace{1em}

\noindent This definition does not take into account the problems associated with categorizing people into such groups or the effect of intersectionality, but these categories function as reference points as they commonly serve as a basis for research, public discourse, and policy decisions.

\subsection{Empirical Example}

To illustrate, we consider a query for which there is no commonly-recognized ``correct'' answer, namely the philosophical and religious question of what does it mean to live a good life. Each culture implicitly or explicitly addresses this through moral, ethical, and religious precepts or arguments, and so the range of possible answers is quite large. Naturally, an LLM asked the question in a general form will respond with the ``greatest hits'' version, focusing on Aristotle, Utilitarianism, Existentialism, etc. This is often what users expect from such a query, in the same way one might expect to cover the `big names' in a one-semester introduction to philosophy class. But we consider a few possible problems with the existing behavior of LLMs. 

First, in accordance with minimal representativeness, as defined above, if asked for a list of philosophers representing different traditions, we desire that any valid response has some non-zero probability of appearing. Secondly, from a task-pragmatic perspective, consider the case of a student or curious individual wishing to learn more about different moral viewpoints. Posing such a question should provide specifics that allow the individual to ask follow-up details or search for more information. Anecdotally, however, there appears to be a tendency to mention specific well-known mainly Western philosophers, while packaging philosophic and religious traditions from the rest of the world into general principles without a specific tradition or person mentioned\footnote{For example, one response in the corpus includes: \\ \emph{The Good life} Philosophers such as Aristotle have proposed that living a good life involves cultivating virtues, engaging in meaningful activities, and developing one's potential to achieve eudaimonia (flourishing). [...] \\ \emph{Relationships}: Human well-being is often seen as intertwined with the quality of one's relationships with others. Social connections, love, and belonging are considered important for overall well-being.} This makes it difficult to pursue more details on these perspectives, and at the same time makes such follow-up seem less likely to be fruitful because it appears to be ``common-sense'' and banal.

Finally, and relatedly, we could consider the extent to which the responses are representative of different cultures or traditions. For example, in the corpus overall (described below) there are 392 mentions of ``Martin Seligman'', an American psychologist who has written on happiness and well-being, while only 62 for ``Ghazali'' (Al-Ghazali), and 52 for ``Farabi'' (Al-Farabi) two of the most influential Islamic philosophers of all time. Again, whether that should be considered culturally-biased or not may depend on specific use-cases, as some users might prefer a temporal focus on currently-living authors, for example. We argue that at a minimum, however, it is worth trying to measure the representativeness among diverse cultural traditions but also the tendency to mention only a narrow range of individuals even when specifically asked for a diverse list.

We generate a corpus of responses from four models four models (GPT-3.5-turbo, Claude-3-sonnet, Gemini-pro, and Llama2-70b), with a series of five different prompts, in order to explore possible differences between models and provide more robust analysis, but this is not intended as a general purpose benchmark.\footnote{For example, temperature=1.0 was used for all models and no GPT-3.5-turbo was provided with a unique seed for each query, an option not available for the other models, although no texts were exact duplicates.}

Without engaging in an exhaustive search of different prompts, we included five prompts with increasing effort focused on generating diverse responses.  We report the prompts used in Table~\ref{tab:prompts}. These begin with a simple query asking in a straightforward manner, then add the word ``diverse'' in prompt 2, then ask for ``as many diverse... as possible.'' While most of the responses to these initial prompts include lists, prompts 4 and 5 specifically ask for the LLM for a list of twenty responses in order to encourage the idea of going beyond the most common responses. Finally, prompt five asks for the list to be generated based on one of a list of 34 regions (``Islamic world'', ``China'', etc.), consistent with the literature on data generation suggesting that attributes can increase diversity \citep{yu_large_2023}.

\begin{table}[ht]
\centering \footnotesize
\caption{Prompt versions}
\label{tab:prompts}
\begin{tabularx}{\textwidth}{lX}
\toprule
Version & Prompt \\
\midrule
v1 & What does the well-being of a human depend on according to philosophers? \\
v2 & What does the well-being of a human depend on according to diverse philosophers? \\
v3 & What does the well-being of a human depend on? Provide as many diverse philosophers and philosophic traditions as possible. \\
v4 & Suggest philosophical perspectives, traditions or schools of thought on the question: \{question as v1 \} reporting a list of twenty names of philosophers who address the question. \\
v5 & Create a list of philosophical perspectives, traditions or schools of thought from \{attribute\} on the question:  \{question as v1 \} reporting a list of up to twenty names of philosophers who directly address the question. \\
\bottomrule
\end{tabularx}
\end{table}

The entities from each response for prompts 1-3 were extracted by asking GPT-3.5-turbo to extract a list of entities from the text, while for prompts 4 and 5 these were parsed directly from the results. One specific challenge is to map all outputs to a unique but comprehensive list of entities. Our approach consisted of first generating a large set of candidate entities, then doing entity resolution using embeddings and DBSCAN algorithm \citep{schubert2017dbscan}, as follows. First, the corpus of possible entities came from (1) the list of entities generated by the LLMs, and (2) extracting entities from the Internet Encyclopedia of Philosophy \citep{iep2024} and then identifying those which are relevant to the question based on a query to GPT-3.5-turbo. An additional pass through the data asking GPT-3.5-turbo for the associated regions was the basis for the region entities mentioned above.  The embedding for each entity was generated using the HuggingFace `all-MiniLM-L6-v2' model, and similar identities identified using DBSCAN with $\epsilon=3.0$. Identities which were not matched to a cluster based on this cutoff were discarded, meaning that there is a possibly still-longer-tail of answers which is not measured by this method. This generated the final `reference' list of 2,693 entities, to which any entity identified in an output was mapped, again using the same embeddings model.

Each step of this process introduces possible errors, in that (a) the the reference list does not include all possible entities, and is likely biased against less mainstream philosophers,\footnote{Even the Internet Encyclopedia of Philosophy, while including diverse philosophy from all over the world, includes a clear focus on Anglo-American and European philosophy, reflecting the interest of its writers and editors.}  (b) the reference list may contain some entities that are not clearly good answers to the question (for example, being philosophers who did not make significant contributions to ethics/ morality). Finally, (c) biased could be introduced by the mapping of entities to the reference list, for example because of multiple names used for the same entity\footnote{For example, the full name for ``Avicenna'' is actually ``Ab\={u} \'{A}l\={\i} al-\d{H}usayn bin \'{A}bdull\={a}h ibn al-\d{H}asan bin \'{A}l\={\i} bin S\={\i}n\={a} al-Balkh\={\i} al-Bukh\={a}r\={\i} '' and is also written ``Ibn Sina'',``Sharaf al-Mulk'', etc. } or names that are close in the embedding space but nonetheless distinct entities. For the comparative purposes of the task, however, it is no clear reason to expect that these biases systematically distort the outcomes.

\subsection{Empirical results}

We analyze the results through three approaches. First, we calculate Shannon's  Diversity Index ($H'$) a version of Shannon entropy based on the natural logarithm that is commonly used in ecological studies to measure diversity within a community:
\[
H' = -\sum_{i=1}^{R} p_i \ln(p_i)
\]
where $R$ represents the total number of unique entities in the dataset (defined by the 2,693 reference entities), and $p_i$ is the proportion of the total population for entity $i$. We also calculate Pielou's Evenness Index ($J'$), which normalizes Shannon's diversity index relative to a maximum possible $H'$ that would occur if all entities were equally-well represented \citep{pielou1966measurement}, corresponding to Definition 1 above.

\begin{figure}[ht]
  \begin{center} 
        \includegraphics[width=0.9\textwidth]{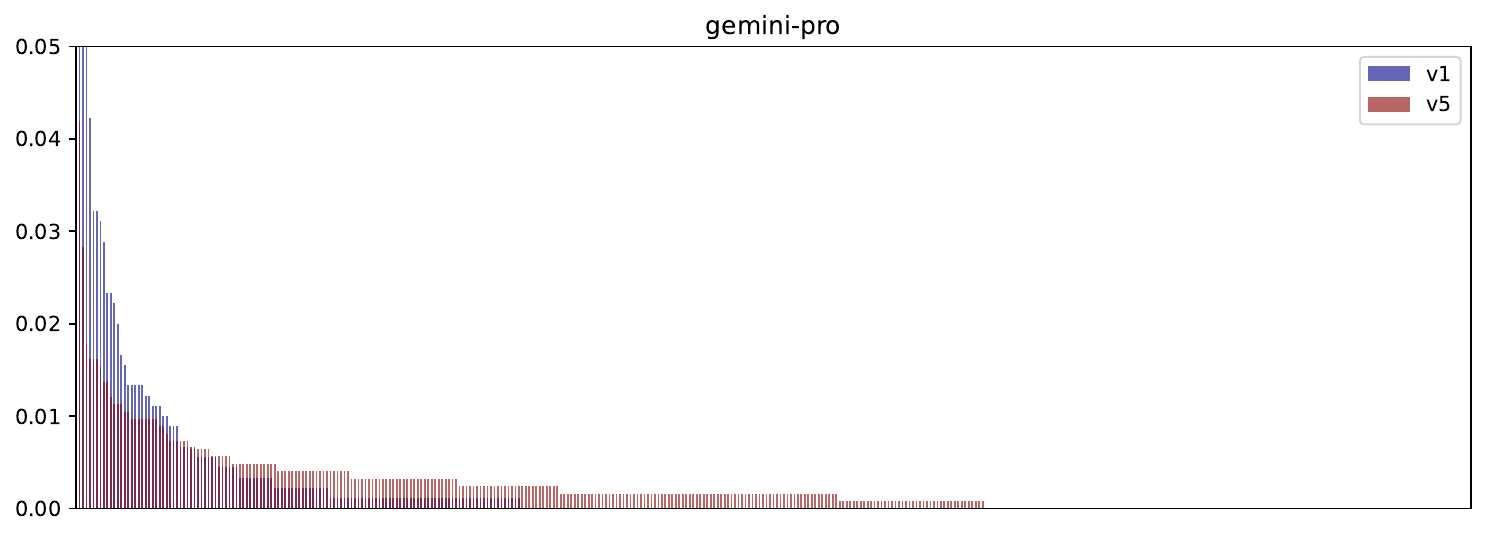}  
        \includegraphics[width=0.9\textwidth]{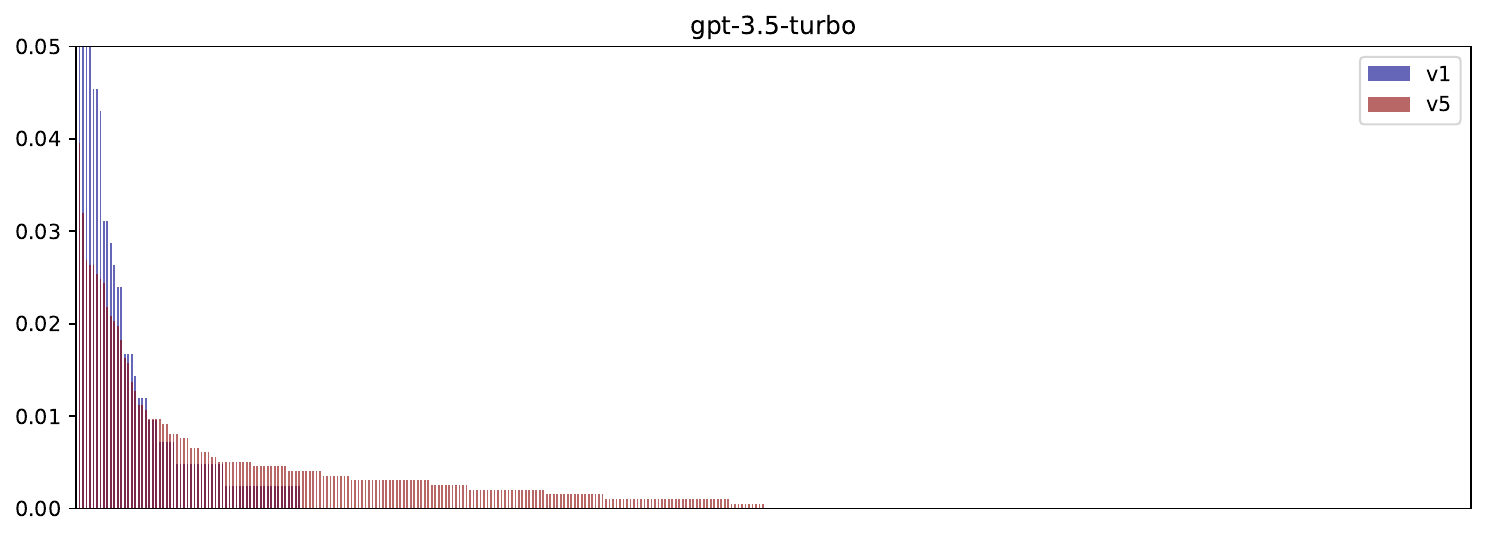}
         \includegraphics[width=0.9\textwidth]{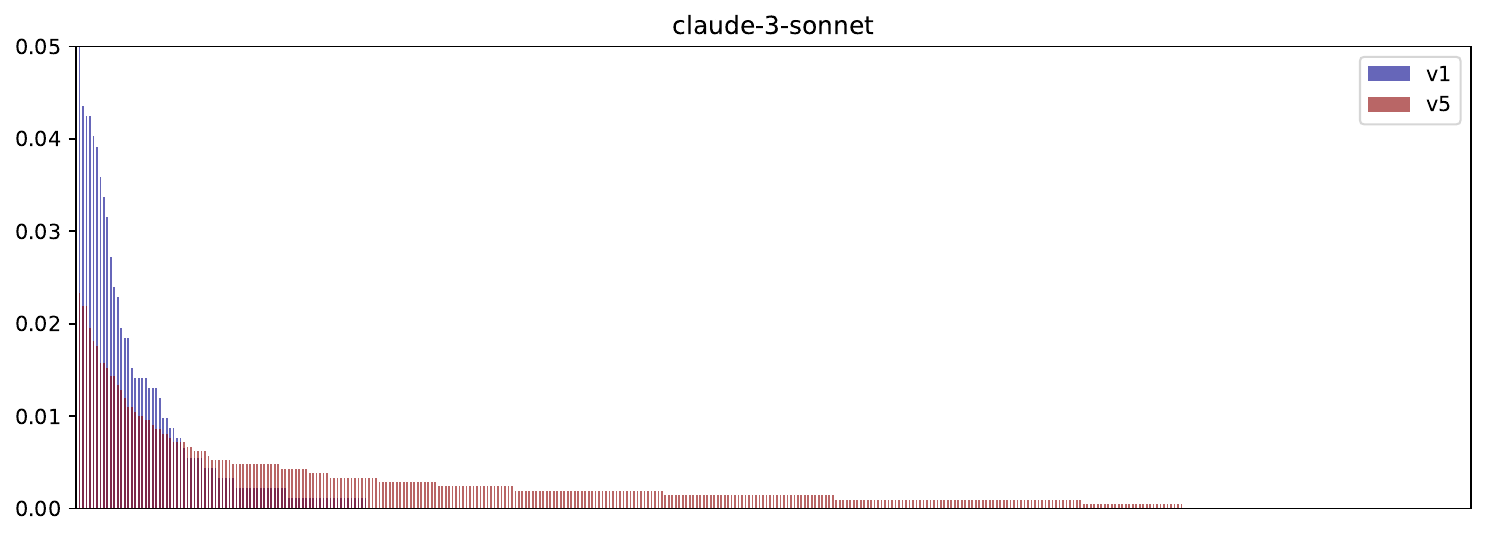}  %
  \end{center}
  \caption{Frequency of philosophical entities based on prompts 1 vs. 5, where (1) simply asks for a response, while (5) asks for a list of twenty individuals or schools of thought from a specific named region. Note: the x-axis is truncated to the 600 most frequent responses, out of 2,693 entities identified in the corpus.}
  \label{fig:freq1v5}
     \vspace{10pt}
\end{figure}

\begin{figure}[ht]
  \centering
  \begin{minipage}{\textwidth}
    \centering
    \includegraphics[width=0.8\textwidth]{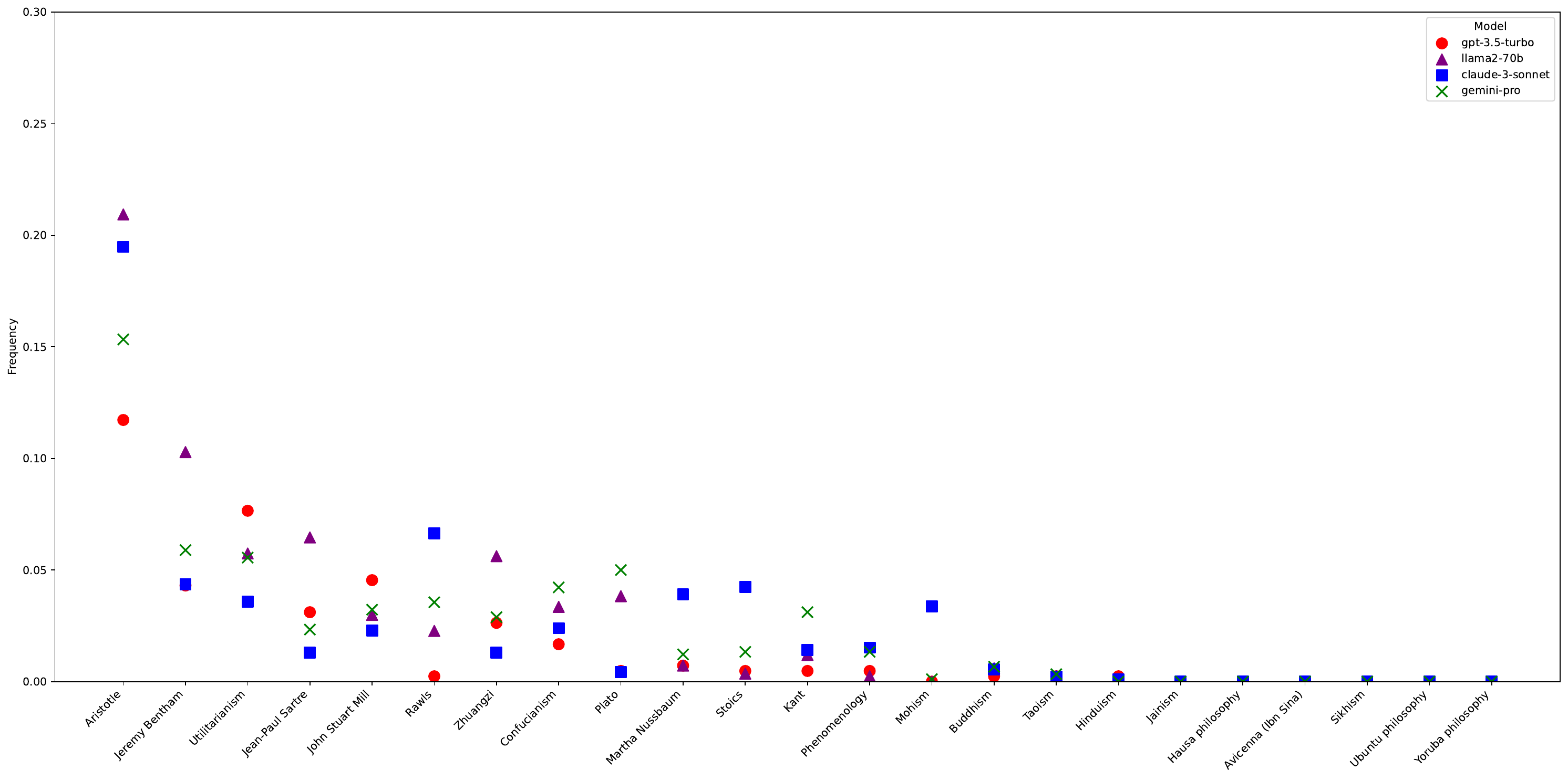}
    \begin{picture}(0,0)
      \put(-110,150){\textbf{Prompt v1}}
         \end{picture}
  \end{minipage}

  \begin{minipage}{\textwidth}
    \centering
    \includegraphics[width=0.8\textwidth]{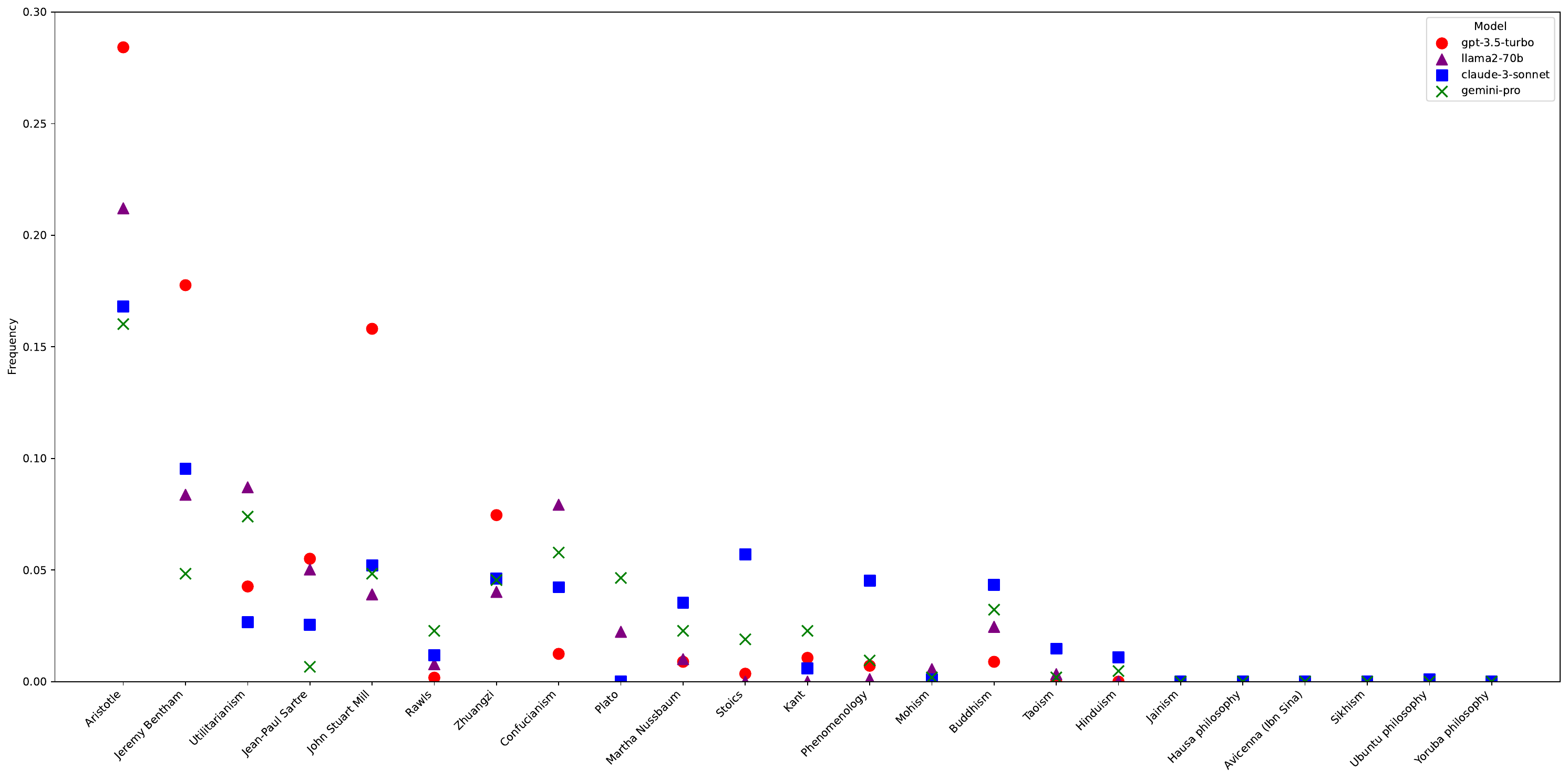}
    \begin{picture}(0,0)
      \put(-110,150){\textbf{Prompt v2}}  
    \end{picture}
  \end{minipage}

  \begin{minipage}{\textwidth}
    \centering
    \includegraphics[width=0.8\textwidth]{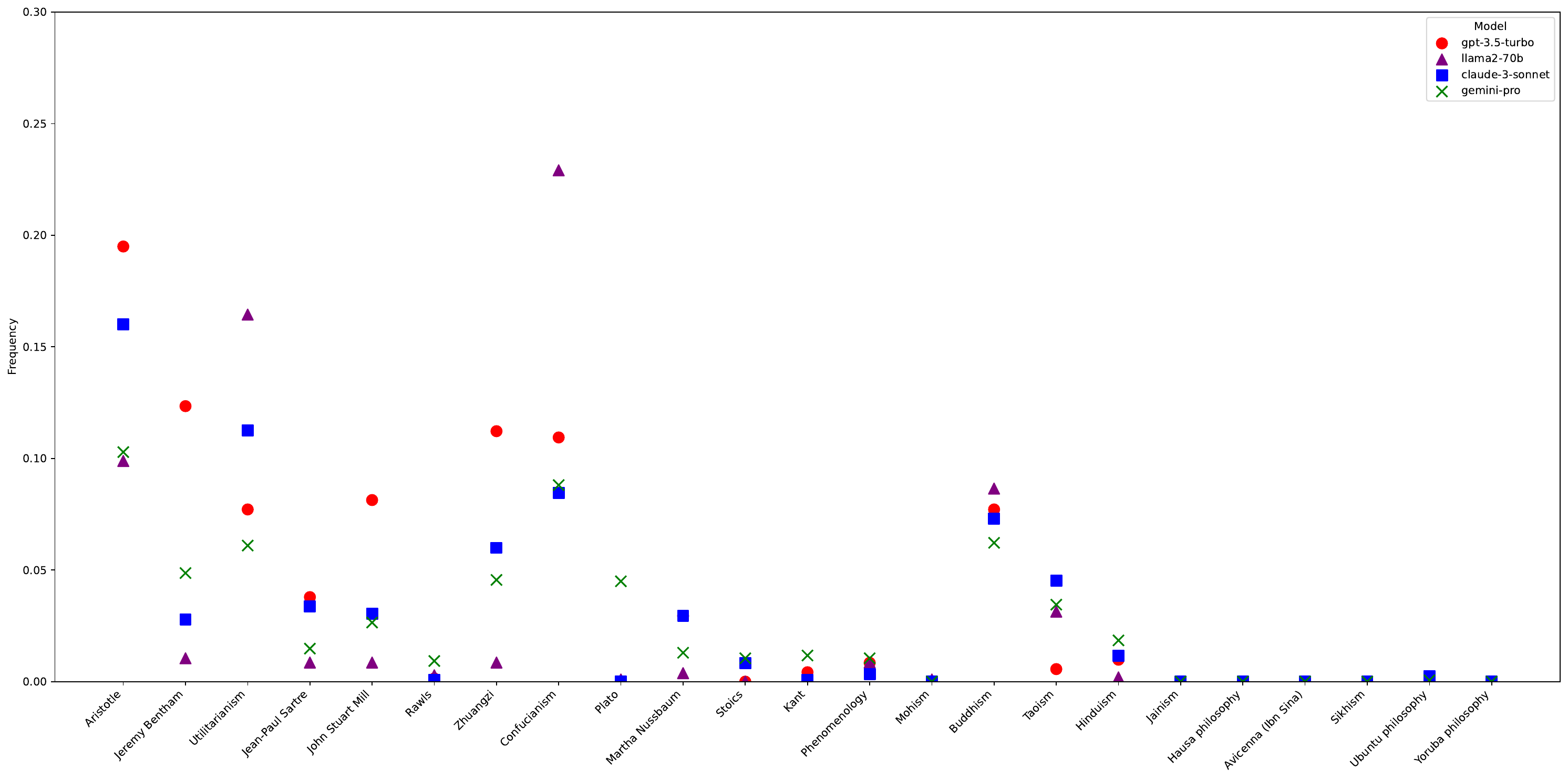}
    \begin{picture}(0,0)
      \put(-110,150){\textbf{Prompt v3}} 
    \end{picture}
  \end{minipage}

  \caption{Frequency for selected mentions}
  \label{fig:freq_sel}
\end{figure}

\begin{figure}[ht]
  \centering
  \begin{minipage}{\textwidth}
    \centering
    \includegraphics[width=0.8\textwidth]{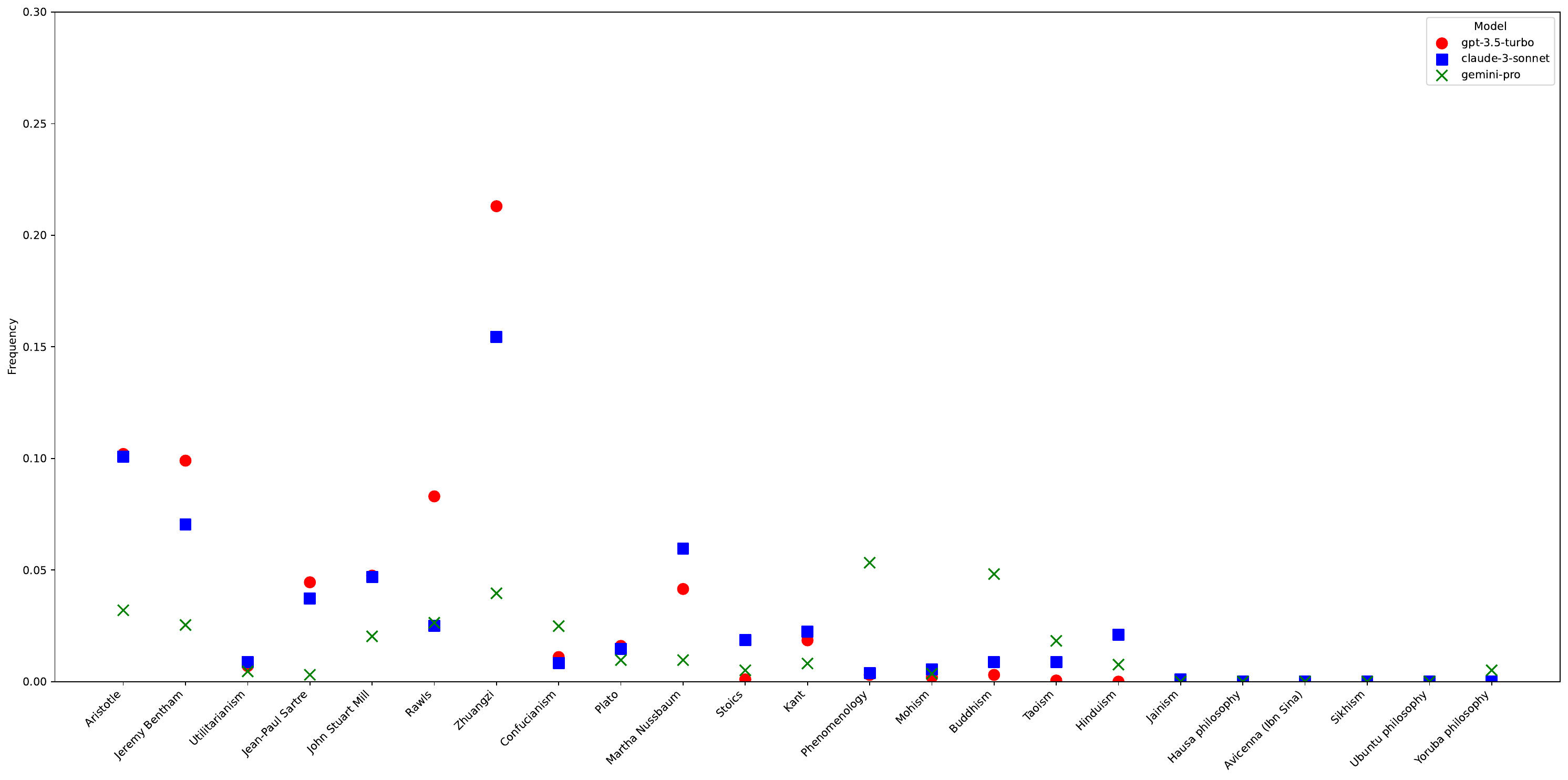}
    \begin{picture}(0,0)
      \put(-110,150){\textbf{Prompt v4}}
         \end{picture}
  \end{minipage}

  \begin{minipage}{\textwidth}
    \centering
    \includegraphics[width=0.8\textwidth]{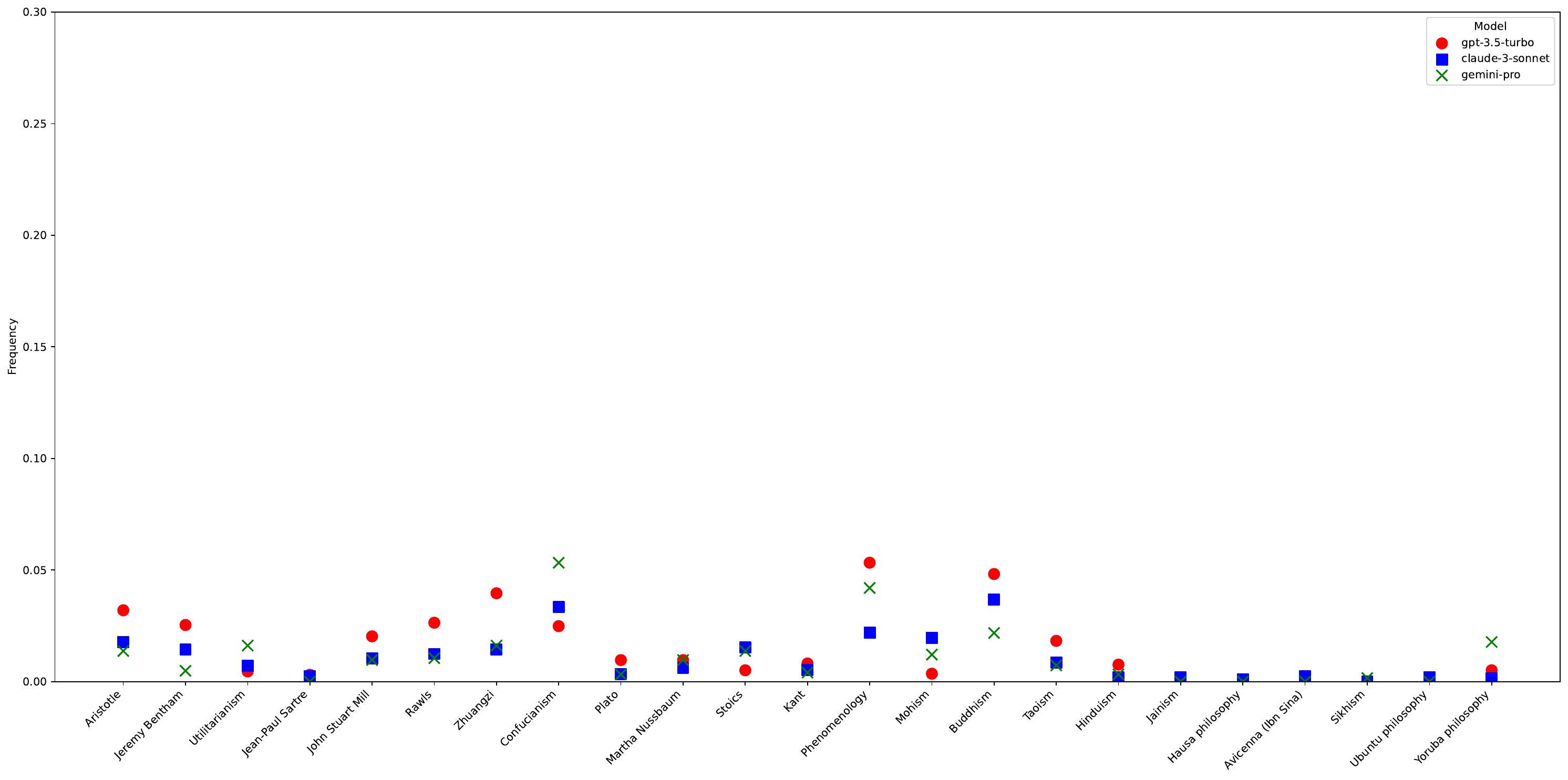}
    \begin{picture}(0,0)
      \put(-110,150){\textbf{Prompt v5}}  
    \end{picture}
  \end{minipage}

  \caption{Frequency for selected mentions. Because prompt v5 encourages greater diversity, there is no one individual with high frequency. }
  \label{fig:freq_sel2}
\end{figure}
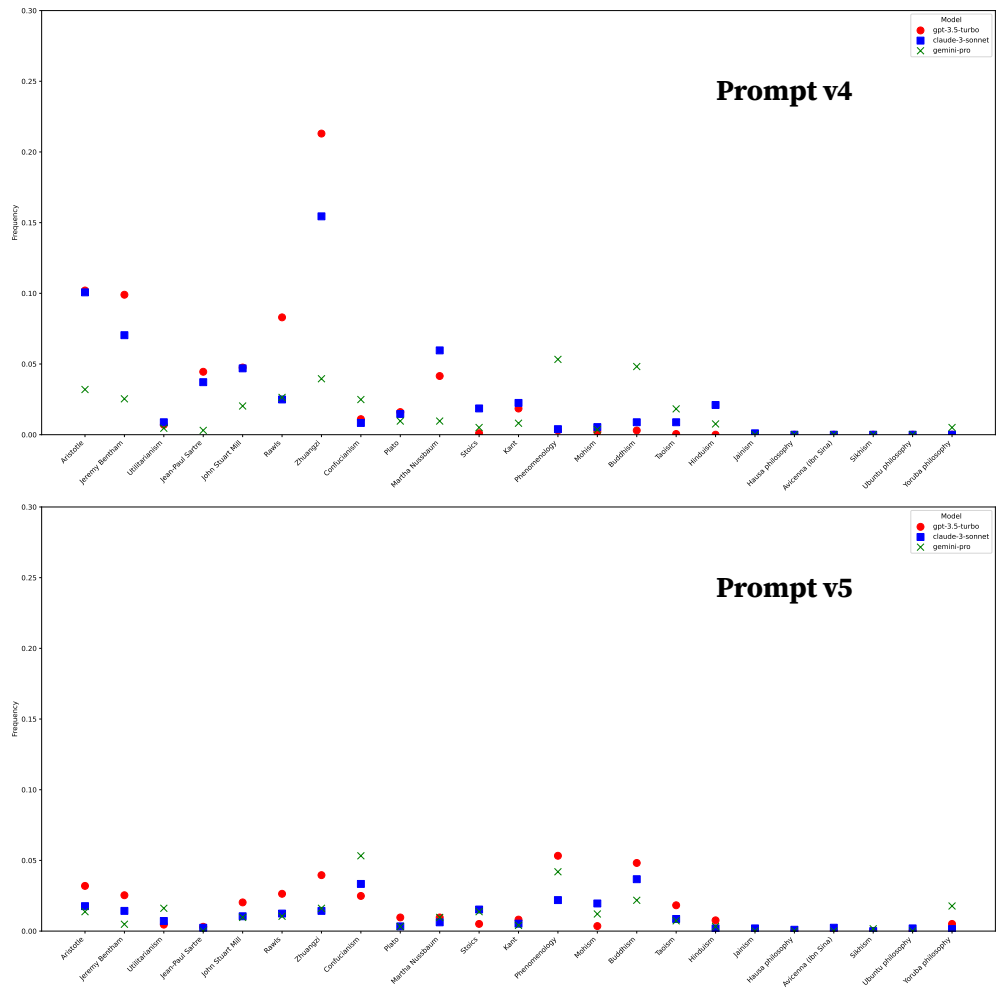

The results are reported in Table~\ref{tab:indices}. There does not appear to be significant variation in the different prompts, with the exception of prompt 5, which does generate significantly more diverse responses (mean $J'$ of $0.86$ compared to $0.55$ for prompt 1). This is represented visually in Figure~\ref{fig:freq1v5}), in which prompt version 5, which requests lists from specific regions, clearly includes fewer references of the most ``popular'' responses and is more likely to include long-tail responses. Finally, to provide specific examples of the output for a set of identifiable entities, Figures~\ref{fig:freq_sel} and \ref{fig:freq_sel2} present the frequencies for the top 10 most popular entities and a few selected other examples from the rest of the distribution. While Aristotle makes up as much as 20\% of all entities identified for Prompts 1 and 2, he is mentioned only half as much with versions 4 and 5. Correspondingly, other responses such as Buddhism and Confucianism is more frequent with the prompts asking for diversity, while more rare entities like Yoruba philosophy or Avicenna are almost never mentioned, regardless of the prompt. The plots also underestimate the full list of 2,693 entities, as they are truncated to the 600 most frequent responses.

\begin{table}[ht]
\centering  \footnotesize
\caption{Shannon and Pielou's Indices by Model and Prompt}
\label{tab:indices}
\begin{tabular}{llrr}
\toprule
    prompt  &    model &  Shannon &  Pielou \\
\midrule
               v1 & claude-3-sonnet &     4.01 &     0.51 \\
               v2 & claude-3-sonnet &     3.78 &     0.48 \\
               v3 & claude-3-sonnet &     3.70 &     0.47 \\
               v4  & claude-3-sonnet &     4.47 &     0.57 \\
               v5  & claude-3-sonnet &     7.02 &     0.89 \\
               v1 &      gemini-pro &     4.60 &     0.58 \\
               v2  &      gemini-pro &     4.23 &     0.54 \\
               v3 &      gemini-pro &     4.54 &     0.57 \\
v5  &      gemini-pro &     6.61 &     0.84 \\
               v1 &   gpt-3.5-turbo &     4.34 &     0.55 \\
               v2  &   gpt-3.5-turbo &     2.94 &     0.37 \\
               v3 &   gpt-3.5-turbo &     3.18 &     0.40 \\
               v4 &   gpt-3.5-turbo &     3.95 &     0.50 \\
v5 &   gpt-3.5-turbo &     6.67 &     0.84 \\
\bottomrule
\end{tabular}
\end{table}

\subsection{Discussion}

The results suggest that prompting strategy is one approach that can, to an extent, improve the diversity of outputs.  But in the corpus including all five prompts, Avicenna or Ibn Sadi appears 57 and 26 times respectively, for 83 total mentions, versus 1,080 mentions for Martha Nussbaum, and 4,779 for Aristotle.\footnote{Admittedly, Aristotle is a good answer, given his specific mention of ``eudaimonia'' that is often phrased in terms of the good life, influence on the subsequent Western tradition, etc. But for reference, the corpus has only 30,315 uses of the article ``the'', so `Aristot' is mentioned once for every 7 times `the' is used, despite most prompts asking for diversity. } The concept of `Ubuntu', which has become relatively well-known among the general population for a non-western philosophical notion, receives 126 mentions.

The limitations of the approach urge caution when making specific claims of cultural bias, but we think that further investigation is merited to better understand not only problems of cultural representativeness but of the general issue of `homogenization' or bias towards popular or common responses identified in the model. One caveat is that the specific wording of the prompts, such as the phrases ``well-being'' and ``philosopher'', might bias the results against inclusion of more diverse ethical traditions grounded in religions or less associated with the specific phrase. Furthermore, additional work could investigate the impact of alternative prompting approaches, different temperature settings and other decoding strategies \citep{wiher_decoding_2022}. This specific philosophic example is obviously limited as well, and could be extended to other domains.

\section{Conclusion}

We provide a theoretical framework for defining ``knowledge collapse'', whereby dependence on generative AI such as large language models may lead to a reduction in the long-tails of knowledge. Our simulation study suggests that such harm can be mitigated to the extent that (a) we are aware of the of the possible value of niche, specialized and eccentric perspectives that may be neglected by AI-generated data and continue to seek them out, (b) AI-systems are not recursively interdependent, as occurs if they use other AI-generated content as inputs or suffer from other generational effects, and (c) AI-generated content is as representative as possible of the full distribution of knowledge.

Each of these suggest practical implications for how to manage AI adoption. First, while our work does not justify an outright ban, measures should be put in place to ensure safeguards against widespread or complete reliance on AI models. For every hundred people who read a one-paragraph summary of a book, there should be a human somewhere who takes the time to sit down and read it, in hopes that she can then provide feedback on distortions or simplifications introduced elsewhere. One extension to the model would be to allow for generational change but endogenize the choice of public subsidies to protect `tail' knowledge. This is arguably what is done by governments that support academic and artistic endeavors that would otherwise have been under-provided by the private market. Protecting the diversity of information means also paying attention to the effect of AI adoption on the revenue streams of journalists that produce and not merely transmit information \citep{cage2016saving}. 

Secondly, there is an obvious need to avoid building recursively dependent AI systems (e.g. where one LLM or agent provides answers based on another AI-generated summary, etc.) and thereby playing an LLM-mediated game of `telephone'. At a minimum, this requires a concerted effort to distinguish human- from AI-generated data. Preserving access to `unmediated' texts, such as through a well-conceived retrieval augmented generation approach, can preserve the long-tails of knowledge \citep{delile_graph-based_2024}, as may generating multiple results and re-ranking \citep{li_search_2023}.

Finally, while much recent attention has been on the problem of LLMs misleadingly presenting fiction as fact (hallucination), this may be less of an issue than the problem of \emph{representativeness} across a distribution of possible responses. Hallucination of verifiable, concrete facts is often easy to correct for. Yet many real world questions do not have well-defined, verifiably true and false answers. If a user asks, for example, ``What causes inflation?" and a LLM answers ``monetary policy'', the problem isn't one of hallucination, but of the failure to reflect the full-distribution of possible answers to the question, or at least provide an overview of the main schools of economic thought.

This could be considered in the setup of frameworks for reinforcement learning from human feedback and related approaches to shaping model outputs, since humans may by default prefer simple, monolithic answers over those that represent the diversity of perspectives. Particular care should also be given in the context of the use of AI in education, to ensure students consider not only the veracity of AI-generated answers but also their \emph{variance}, \emph{representativeness}, and \emph{biases}, that is, to what extent they represent the full distribution of possible answers to a question.

The scaling laws \citep{hoffmann2022training} demonstrate the advantage of training LLMs on the maximum amount of (quality) data. A valuable empirical question is therefore whether this leads to increasing or decreasing diversity within the training data (and the raises the related problem of the lack of transparency in the data used to train models). There are many diverse texts that could be included to expand the corpus, but practically, the approach of market-focused participants may be to focus on seeking texts with the lowest marginal cost (conditional on quality). This might exacerbate a reliance on texts that are not representative of the general public, such as if social media texts are easy to collect but not representative of the perspective of people who don't have access to social media or self-select out of them. Or, optimistically, companies with a global audience might be incentivized to seek out ``low and very-low resource languages''  \citep{team2023gemini} and perhaps even the viewpoints and cultural perspectives of diverse users. Consideration should be given to ensuring and encouraging such diverse inputs as well as to monitoring of the diversity of outputs. Finally, additional work should look at specifically how the use of AI is integrated into the larger pre-existing practices of social learning \citep{mercier2019utilizing,de2021cultural}.

\clearpage

\vskip 0.2in
\bibliography{collapse}
\bibliographystyle{aaai}

\clearpage

\appendix
\section*{Appendix A}

\vspace{1em}
\subsection{Comparing width of the tails}

\begin{figure}[ht]
  \begin{center}
        \includegraphics[width=0.35\textwidth]{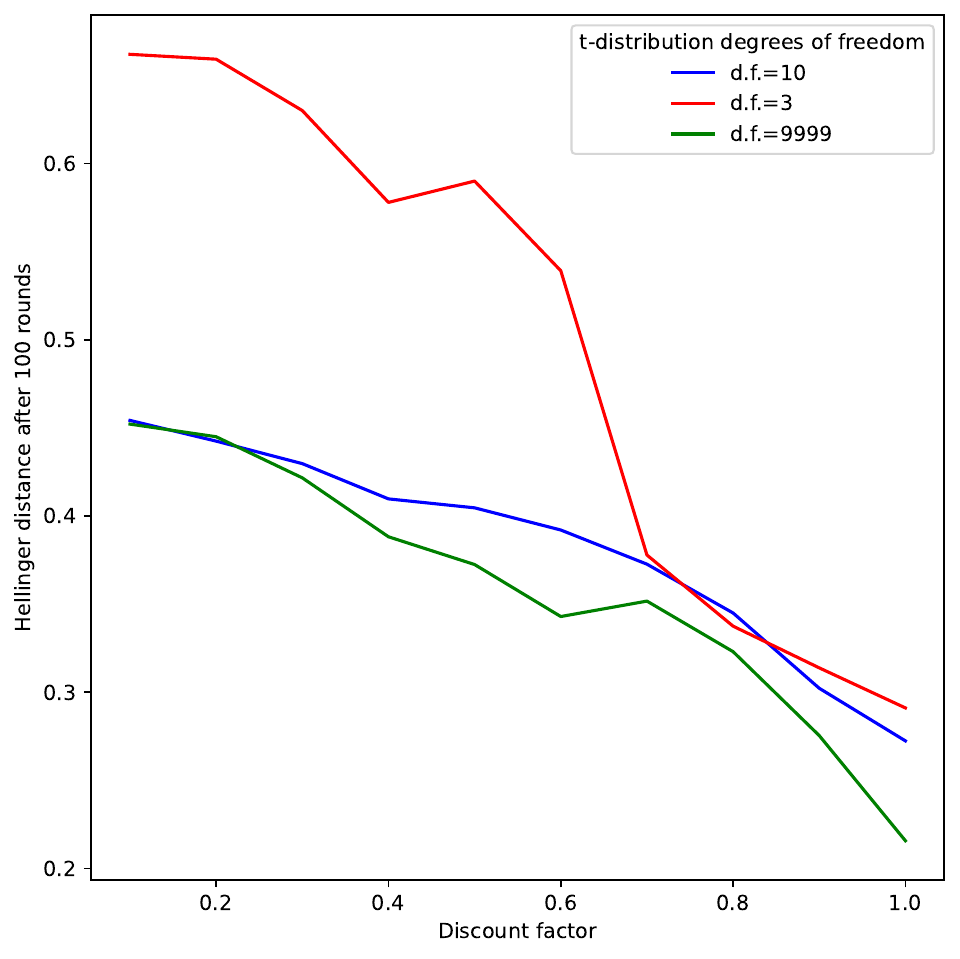}  
  \end{center}
  \caption{Discount rate and varying thickness of the tails}
  \label{fig:tail_thickness}
     \vspace{10pt}
\end{figure}

As mentioned above, the reported results used a t-distribution with 10 degrees of freedom, which has slightly wider tails than a standard normal distribution. We can compare the results with a standard normal distribution (\ie  a t-distribution as the degrees of freedom becomes large) or with wider tails. In Figure~\ref{fig:tail_thickness}, we plot a comparison of the results from the main section (with 10 degrees of freedom with wider or narrower tails (3 and 9999 degrees of freedom respectively). The main difference is for more extreme discounts provided by AI ($ < 0.7$), for which the wider tails contribute to knowledge collapse (\ie generate a public knowledge distribution further from the true distribution).  Narrower tails, such as from a standard normal distribution, generate results broadly similar to the main model.  Thus, as expected more information in the tails makes the effect of knowledge collapse more pronounced, but is plays less of a role than the other parameters discussed above in determining the dynamic of collapse.

\vspace{1em}

\subsection{Defining knowledge collapse} \label{sec:defining_kc}

\vspace{0.5em}

To define knowledge collapse we need to distinguish between a few conceptual sets of `knowledge', whether or not these are empirically observable.\footnote{The broadest definition of `human knowledge' might encompass all the beliefs, information, values, and representations of the world ever held by humans anywhere on earth, whether recorded or not. We are unable to access almost all of this, and we tend to assume that the useful parts of this have been passed on to others, but theoretically we might want to allow for the fact that, for example some human somewhere once had an important, original, and useful belief just before they, say, got hit by a car and could not tell anyone. Secondly, in using the term `knowledge', we do not restrict our focus based on the truth of the beliefs held, such that in referring to `human knowledge' we refer to a variety of beliefs and statements, some of which contract others. }
First, we consider the broad set of historical human knowledge that was at one point held in common within communities of humans, shared and reproduced in a regular way, which we might call \textbf{`broad historical knowledge'}. 

Second, we consider the set of knowledge that is held or accessible to us, (humans who are living in a given epoch), which we call  `\textbf{available current knowledge}.' In the example cited in the main section, the ancient Roman recipe for concrete is part of broad historical knowledge but not part of available current knowledge.

Technological innovations from the printing press to the internet to AI mediate human interactions and human's exposure to historical and current sources of knowledge. The net effect might be to restrict or expand access to diverse knowledge and the long-tails of human knowledge. For example, the digitization of archives might make obscure sources available to a wider audience and thus increase the amount of `broad historical knowledge' that is part of the `available current knowledge.' 

We also distinguish a third, narrower set of knowledge, which reflects not what is theoretically accessible to humans but which is readily part of human patterns of thinking or habits of thought. This we call `human memory knowledge' or `\textbf{human working knowledge}' by reference to human working memory.

For example, consider the problem of listing all the animals that have ever existed on earth. There might be some that humans previously knew about, but which subsequently went extinct and which do not exist anywhere among the scientific literature or individuals currently living on earth. More narrowly, the set of ``available current knowledge'' corresponds to the set of all animals that a team of all biologists could compile with access to the internet and other records. Finally, however, if we were able to conduct a survey of all humans on earth and ask them to name as many animals as possible in, say, one day, we would come up with a more limited list (that would include many repetitions).

In many practical applications, `human working knowledge' is the most relevant because it is the knowledge that shapes human action and reflection. A doctor considering possible sources of a crossover pathogen might rely on their knowledge of common species in asking a patient if they had recently been in the presence of certain animals  (even if a researcher who specializes in this area might consult know more and sources to find a longer possible list). A linguist trying to evaluate or create possible linguistic theories implicitly bases their judgement on the known language families and their structures, and so on. Edison and his team famously tried thousands of different filament materials, but if it bamboo had not been among the materials that came to mind as they searched alternatives, a practical electric bulb may have been invented only later. 

Finally, it is useful to define the `\textbf{epistemic horizon}' as the set of knowledge that a community of humans considers practically \emph{possible to know} and \emph{worth knowing}.\footnote{In economic terms, it is the set of information that for which the individual believes the expected returns are greater than the expected costs. This might be considered for a specific task or set of tasks, but could be generalized to the set of knowledge for which she expects positive gains over a period of time, her lifetime, or for society over a finite or infinite horizon with discounting.} A common controversy in the public imagination is whether traditional medicines are worth consideration when searching for medical cures. Such traditional medicines might be outside of the epistemic horizon because they are not written down in the scientific literature, are only known by individuals speaking lesser known languages, or because the scientists in question consider them too costly to acquire or unlikely to be beneficial. One way to think about this relationship is as a generalization of `availability bias', in which we take the set of readily recalled information to be more likely, important, or relevant \citep{tversky_availability_1973}.

In these terms, we define \textbf{`knowledge collapse'} as the progressive narrowing over time (or over technological representations) of the set of human working knowledge and the current human epistemic horizon relative to the set of broad historical knowledge.

\begin{definition}[\textsc{Knowledge Collapse}]
Consider a historical set of human knowledge $p_{\text{true}}(x)$ represented as a distribution over time or technological representations. We define \textbf{knowledge collapse} as the progressive narrowing of this distribution such that the variance of $p_{\text{true}}(x)_t$, denoted as $\sigma^2(p_{\text{true}}(x)_t)$, approaches zero as time $t$ progresses. Formally, we can represent this as:
\begin{align}
    \lim_{t \to \infty} \sigma^2(p_{\text{true}}(x)_t) = 0
\end{align}
In the limit, the probability density function of $p_{\text{true}}(x)_t$ approaches a Dirac delta function centered at some knowledge point $\mu_t$, i.e.,
\begin{align}
    p_{\text{true}}(x)_t \xrightarrow{\text{d}} \delta(\mu_t)
\end{align}
where $\xrightarrow{\text{d}}$ denotes convergence in distribution.
\end{definition}

On a theoretical level, the idea of epistemic horizon has an intellectual heritage in Immanuel Kant's argument about the forms and categories of understanding that underly the possibility of knowledge \citep{kant1933critique}. Subsequent authors expanded on the implications if these categories are in some way fashioned by one's upbringing and community \citep{hegel2018hegel,mannheim1952sociological}.\footnote{\eg ``If man received every thing from himself and developed it independently of extrinsic objects, then a history of a man might be possible, but not of men in general. But as our specific character resides precisely in this, that, born almost without instinct, we are raised to manhood only by lifelong practice, on which both the perfectibility as well as the corruptibility of our species rests, so it is precisely thereby that the history of mankind is made a whole: that is, a chain of sociability and formative tradition from the first link to the last.\citep{herder2024ideas}} A related concern is the way that the scientific community can be, at least during certain epochs, bounded by its inherited understanding of the world \citep{kuhn1997structure,zamora_bonilla_science_2006}. As noted above, specific technological forms may generate a flood of information that inhibit the communication of information \citep{pfister_logos_2011}.  
 
Finally, one of the challenges presented by the `epistemic horizon' (as of that of an `event horizon'\footnote{Technically, we can observe the `shadow' of an event horizon \citep{khodadi2020black}}) is that we cannot observe directly its limits. For example, the presence of an event our current model takes to be very rare  (\eg a ``20-sigma'' event) can suggest our current model is incorrect, but in the absence of such a rare event, we cannot know if the current tails of knowledge are correct or too thin \citep{taleb_black_2007}. These considerations suggest the concern of generational knowledge collapse is plausible and an unbounded optimism in the ability of rational actors to update on the value of tail knowledge may be shortsighted.

\end{document}